%% file: main.tex
\documentclass[runningheads]{llncs}

\usepackage{longtable} 
\usepackage{amsmath,amssymb,amsfonts}
\usepackage{booktabs}
\usepackage[expansion=true]{microtype}\usepackage{url}
\usepackage{cite}
\usepackage{listings}
\usepackage{graphicx}
\usepackage{tabularx}
\usepackage{stmaryrd}
\usepackage{xcolor}
\usepackage{hyperref}
\usepackage{forest}

\lstdefinestyle{promptstyle}{
  basicstyle=\ttfamily\scriptsize,
  frame=single,
  framerule=0.35pt,
  rulecolor=\color{black!45},
  backgroundcolor=\color{black!2},
  framesep=3pt,
  xleftmargin=2pt,
  xrightmargin=2pt,
  aboveskip=4pt,
  belowskip=4pt,
  breaklines=true,
  breakatwhitespace=true,
  columns=fullflexible,
  keepspaces=true,
  showstringspaces=false,
  lineskip=-1pt
}

\lstnewenvironment{prompt}[1][]
  {\lstset{style=promptstyle,#1}}
  {}

\newcommand{\Compile}{\ensuremath{\mathsf{Compile}}}

\begin{document}

\title{FOL-Proof: Verifying LLM Text-to-Logic Translations via Deterministic Semantics}
\titlerunning{FOL-Proof: Verifying LLM Text-to-Logic Translations}

\author{
    Yisrael Clark\inst{1}, Guy Katz\inst{1}
}

\authorrunning{Y. Clark et al.}

\institute{
    The Hebrew University of Jerusalem,
    Jerusalem, Israel\\
    \email{yisrael.clark@mail.huji.ac.il}
}

\maketitle

\begin{abstract}
  Large Language Models (LLMs) are increasingly used to translate
  natural-language statements into first-order logic (FOL) to improve reasoning. However,
  LLMs are prone to errors, and so a semantically unfaithful FOL formula could corrupt downstream
  reasoning. We present FOL-Proof, a semantic firewall pipeline that
  validates LLM-generated FOL translations against independently
  derived linguistic interpretations. FOL-Proof operates by deterministically obtaining
  a pool of candidate translations via semantic
  analysis tools, such as the English Resource Grammar and Minimal
  Recursion Semantics; then applying an SMT solver to
  check each derived reading for logical equivalence to the LLM-generated formula. If the
  translation is faithful it is accepted; otherwise, it is rejected, or optionally repaired via
  repeated invocations to the LLM. The procedure is 
    conservative: incomplete search, unsupported constructions, or
    inconclusive checks yield abstention rather than false rejection.
  We evaluated FOL-Proof on established benchmarks
  (ProofWriter, RuleTaker and PrOntoQA), and also on a new adversarial benchmark
  (LongFaithCheck).
Our experiments show that 
  FOL-Proof attains near perfect selective accuracy,
  while direct LLM judges and round-trip baselines show substantially worse performance.

\end{abstract}

\section{Introduction}
\label{sec:introduction}

Large Language Models (LLMs) have demonstrated remarkable capabilities
across a wide range of natural language processing tasks, including
machine translation, text summarization and question
answering~\cite{zhu2024multilingual,zhang2024benchmarking,holtermann2024evaluating}. They
have also shown promising results in reasoning tasks, including
mathematical problem solving, code generation, and logical
inference~\cite{guo2025deepseek,zhuo2025bigcodebench,cai2025deductive}.

\begin{figure}[htbp]
    \centering
    \includegraphics[width=1\linewidth]{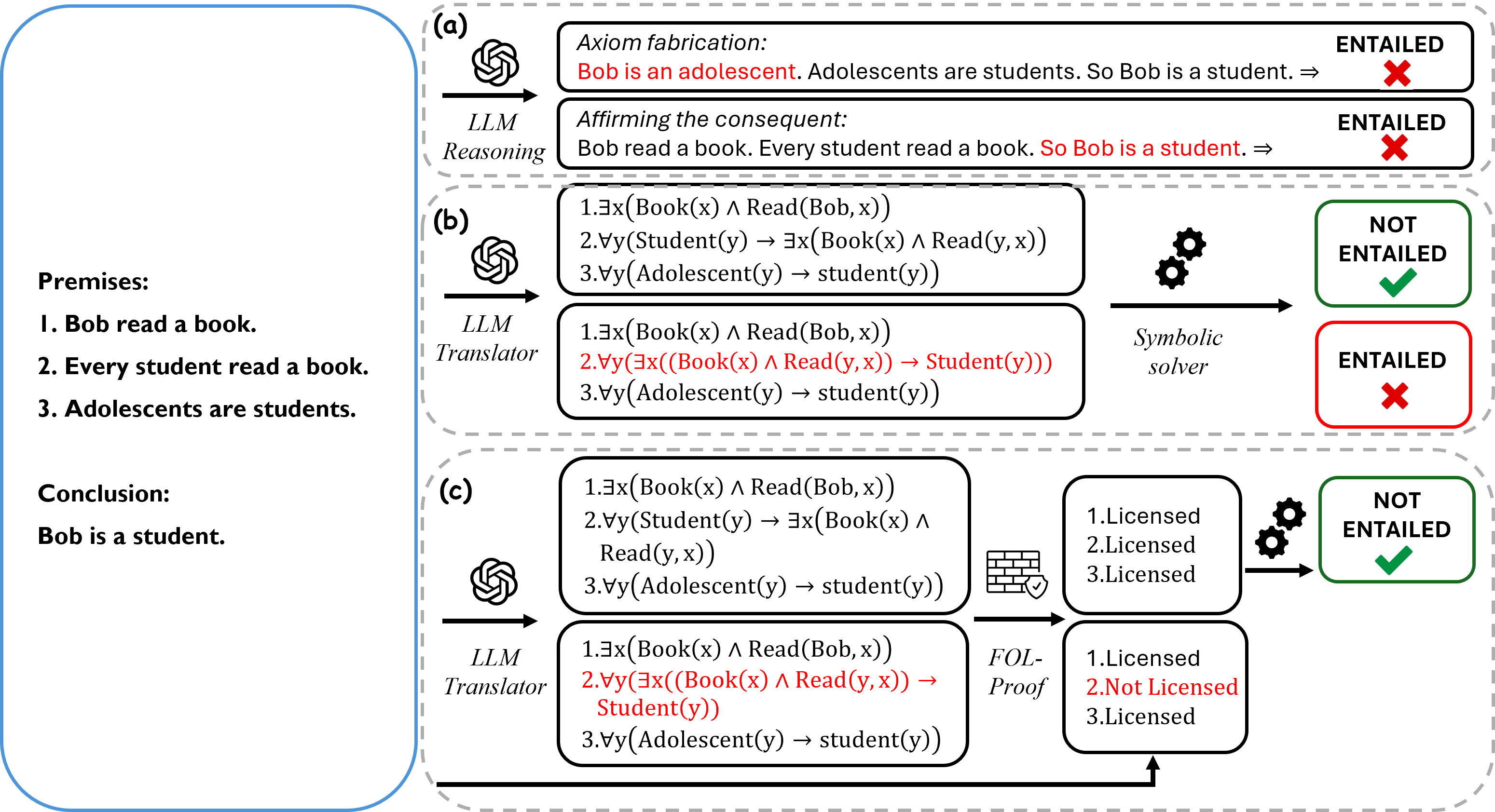} 
    \caption{Reasoning problem example. The correct answer is that the conclusion is not entailed by the premises. (a) LLM reasoning can produce logical fallacies. (b) LLM translators can mistranslate and cause the solver to produce the wrong answer. (c) FOL-Proof validates the translation.}
    \label{fig:reasoning_img}
\end{figure}

Despite these advances, when it comes to deductive tasks (i.e., deciding whether a conclusion is entailed by given premises), LLMs do not guarantee sound reasoning~\cite{zheng2026beyond, pei2025fover,
  deveci2026diagnostic}. 
They are prone to many logical fallacies: fabricating axioms, violating the rules of conditional logic, and employing circular logic~\cite{olausson2023linc, jacovi2024chain, raiyan2026mathematical,
  singh2026clinical,jin2022logical}. 
  For example, consider the reasoning problem in Figure~\ref{fig:reasoning_img}(a). The correct logical answer is that the conclusion is not entailed---Bob is not necessarily a student. But an LLM might conclude otherwise by fabricating an axiom or affirming the consequent.

To address these challenges, the research community has turned to
neuro-symbolic integration~\cite{garcez2019neuralsymbolic,
  creswell2022selection, pan2023logiclm, olausson2023linc}. The core
idea is to use the LLMs for auto-formalization---translating sentences
from natural language to formal logical languages. Then, the generated formulas are offloaded to symbolic solvers for downstream reasoning. This architecture avoids the pitfalls of violating logical reasoning. Neuro-symbolic architectures have demonstrated empirical gains over pure
neural baselines across a spectrum of logical
domains~\cite{szegedy2020promising, wu2022autoformalization,
  bansal2025robustness, putra2026nl2logic, brunello2026struggle}.

However, this pipeline is prone to semantic infidelity\cite{mohammad2026faithfulness,
  brunello2026fixing}. Simply generating a well-structured logical
sentence does not prove that it is faithful to the original,
natural-language sentence; and indeed,
prior work has shown that LLMs frequently mistranslate natural-language statements. Models may select the wrong quantifier or quantifier scope, drop or invert a negation, misuse logical connectives (as is demonstrated in Figure~\ref{fig:reasoning_img}(b)), omit premises or predicates, or even introduce clauses that are not supported by the source text~\cite{wang2026formalrx, thatikonda2024strategies,
kim2026formalizationgaming, mcginness2024automated}. In such cases,
the solver may reach the wrong conclusion, due to being fed incorrect
input.
Therefore, the neuro-symbolic approach does not eliminate the
potential for incorrect reasoning when using LLMs, but rather moves
the problem to the auto-formalization stage.

Existing approaches try to mitigate auto-formalization errors by
guiding the generation process (e.g., prompt engineering, intermediate representations), or through post-hoc LLM judgment and repair. However,  such approaches are susceptible to the LLM's
inherent statistical nature---be it at the generation stage or the verification stage---and fail to provide a certification of
correctness.

Yet another challenge is posed by the inherent ambiguity of natural
languages. Recent human audits of auto-formalization corpora reveal
that up to 48\% of sentences exhibit legitimate linguistic
multiplicity that cannot be reduced to a single canonical reference
formula~\cite{brunello2026fixing}.  For example, ``Every student read
a book'' can be translated into first-order logic (FOL) as
$\forall y\,\bigl(\mathit{Student}(y) \rightarrow \exists
x\,(\mathit{Book}(x)\land\mathit{Read}(y,x))\bigr)$ (Every student
read a book, but not necessarily the same one) or as
$\exists x\,\bigl(\mathit{Book}(x)\land \forall
y\,(\mathit{Student}(y)\rightarrow\mathit{Read}(y,x))\bigr)$ (There
exists a book that every student has read). Both are legitimate, due
to the inherent ambiguity of the sentence.  Consequently, methods that evaluate
auto-formalization via comparison with single ground truth translation are inherently inaccurate~\cite{liu2025gted,mohammad2026faithfulness}.

Here, we propose a novel approach: derive a set of formal
interpretations from an independent, linguistically grounded semantic
analysis of the source sentence; and then use these interpretations as
references against which a proposed formalization is checked. Checking
is performed by evaluating logical equivalence using an SMT solver.
We hypothesize that this checking can retain only high-fidelity
translations for downstream reasoning, filtering out formalizations
that are not licensed by the linguistic analysis before they reach the
solver, thus improving logical reasoning (see
Figure~\ref{fig:reasoning_img}(c)).  
We term this technique 
\emph{FOL-Proof}.

In order to improve the soundness of our approach, we deliberately
designed FOL-Proof to be conservative. It certifies a candidate when some
deterministically derived reading is equivalent to it, and rejects it conclusively only when
the search over parses and scope readings is provably complete and every
adaptable reading is non-equivalent. In all other cases (incomplete search,
unsupported constructions, inconclusive equivalence checks) it abstains, so
resource bounds are never mistaken for semantic
disagreement.

For our implementation of FOL-Proof, we combine the English Resource
Grammar (ERG)~\cite{flickinger2000erg,flickinger2014erg}, Minimal
Recursion Semantics (MRS)~\cite{copestake2005mrs}, and a scope
resolution tool~\cite{kollerthater2005} to generate the pool of
candidate translations. We use a deterministic MRS-to-FOL adapter to
translate these references into formulas, and apply the Z3 SMT
solver~\cite{demoura2008z3} to check equivalence between them and the
LLM-based translation being considered. MRS is useful
for this purpose because it can represent semantic structure while
leaving certain scope choices underspecified; these alternatives can
subsequently be resolved rather than prematurely be collapsed into a
single logical formula, thus preserving natural language ambiguities. 

Our evaluation shows that FOL-Proof provides highly reliable semantic
validation across three well-established reasoning benchmarks with
thousands of items, as well as on a new benchmark that we
present here, consisting of adversarial instances. We evaluate
FOL-Proof on controlled mutations and LLM translations, and find that
it achieves near perfect accuracy across tens of thousands of
formalizations, outperforming direct LLM judges and
round-trip baselines. In addition, it accounts for ambiguity and accepts multiple valid translations.
When used as a gate for downstream reasoning,
FOL-Proof attains 100\% selective accuracy on all answered
problems. Moreover, incorporating FOL-Proof into an
iterative repair loop improves repair rates from 4.8\% to 22.8\% when providing feedback, and increasing to 62.6\% when exposing the closest licensed reading.

To summarize, this paper makes the following contributions:
\begin{enumerate}
  \item We propose a novel approach to verifying LLM auto-formalizations into FOL by certifying them against deterministic English parses. 
  \item We develop FOL-Proof, an implementation of the above approach.
  \item We introduce a new benchmark designed to challenge LLMs' auto-formalization and judging capabilities.
  \item We evaluate FOL-Proof on controlled semantic mutations, LLM translations, end-to-end reasoning and a repair loop, and compare it with direct LLM judges and round-trip validators on ProofWriter~\cite{tafjord2021proofwriter}, PrOntoQA~\cite{saparov2023prontoqa}, RuleTaker~\cite{clark2020ruletaker}, and our new benchmark. We find that our approach can establish near perfect accuracy when it does not abstain. 

\end{enumerate}

The rest of this paper is organized as follows.
Section~\ref{sec:related} reviews related work, and
Section~\ref{sec:background} introduces the necessary background
and formalizes the semantic verification problem.
Section~\ref{sec:architecture} presents the FOL-Proof architecture
and verification procedure, while
Section~\ref{sec:implementation} describes its implementation.
Section~\ref{sec:evaluation} presents our experimental evaluation,
and Section~\ref{sec:conclusion} concludes and discusses future work.

\section{Related Work}
\label{sec:related}

\paragraph{Neuro-symbolic reasoning.} Neuro-symbolic
architectures~\cite{pan2023logiclm, olausson2023linc,feng2026vericot}
have repeatedly demonstrated superior performance over those of pure
neural baselines, by offloading formal execution to symbolic reasoning
engines. Reasoning engines successfully
utilized in this way include automated theorem provers (ATPs) and
Satisfiability Modulo Theories (SMT) solvers, such as
Z3~\cite{demoura2008z3}, Vampire~\cite{kovacs2013first},
Prover9~\cite{mccune2005prover9}, and Lean-4~\cite{demoura2021lean4}.
These results span a spectrum of domains---ranging from FOL reasoning (\text{e.g.}, \textsf{FOLIO}~\cite{han2024folio},
\textsf{ProverQA}~\cite{qi2025proverqa},
\textsf{LogicSkills}~\cite{rabern2026logicskills}) and multi-hop rule
deduction (\text{e.g.},
\textsf{ProofWriter}~\cite{tafjord2021proofwriter},
\textsf{PrOntoQA}~\cite{saparov2023prontoqa},
\textsf{LogicNLI}~\cite{tian2021logicnli}) to complex
constraint-satisfaction and analytical puzzles (\text{e.g.},
\textsf{AR-LSAT}~\cite{zhong2022arlsat},
\textsf{ZebraLogic}~\cite{lin2025zebralogic},
\textsf{SATBench}~\cite{wei2025satbench}). In our work, we focus on FOL reasoning and employ SMT solvers.

\paragraph{Reference-based equivalence and structural metrics.}
Given a ground-truth (GT) formula (i.e., a correct
  translation, used as a point of reference), there exist different
  approaches for comparing the candidate translation against it,
  typically tailored to the domain of the formula in question.
  String metrics such as BLEU are brittle under renaming
and reordering~\cite{thatikonda2025assessing, wang2026formalrx}. Solver-based approaches instead decide equivalence by
refuting $\neg(\phi\leftrightarrow\psi)$ with an SMT solver or
ATP~\cite{murphy2024autoformalizing,singh2026verge}. Other methods exist for comparing Lean statements~\cite{liu2025rethinking} or temporal logic~\cite{quansah2026neuronnl2ltl}. When a metric is needed beyond simple equivalence, some methods employ graph distance calculations~\cite{opitz2023smatch, liu2025gted}.
These methods are exact or graded with
respect to the GT, but are also limited with respect to that GT: they cannot account for ambiguity, and are inapplicable
when no GT is provided.
In FOL-Proof we also use a negated bi-implication check (Sect.~\ref{sec:verdicts}),
but do not assume we are provided with a GT; instead, we use a set of 
grammar-licensed readings.

\paragraph{LLM-mediated verification and selection.}
When no GT reference exists, several systems perform a roundtrip translation---converting the
translation back into natural language and comparing it with the source. VeriTrans uses a
thresholded round-trip similarity score as an acceptance
gate~\cite{kodakandla2026veritrans}, Monty scores bidirectional
clausal coverage with an LLM judge~\cite{liu2026monty},
\(\forall\)uto\(\exists\)val checks truth preservation over repeated
formal-to-natural-language cycles with formal
verifiers~\cite{karia2024autoeval}, ProofBridge combines Lean type
checking, proof repair, and bidirectional theorem-equivalence
checks~\cite{jana2025proofbridge}, and VeriCoT and VERGE use LLM-based
and solver-assisted consistency
checks~\cite{feng2026vericot,singh2026verge}.  These methods can be
used without a GT, but the semantic evidence they provide remains
limited by the quality of the learned reconstruction step: a
systematic error in the forward translation can be preserved by the
backward one, and lexical or embedding similarity does not entail
semantic equivalence~\cite{li2026stable, mohammad2026faithfulness}.

An alternative approach employs LLM-as-a-judge to score the fidelity of translations, while incorporating various other components for improving the accuracy of the translation~\cite{ ryu2025clover,thatikonda2025assessing,brunello2026fixing}. For example, CLOVER uses a SAT solver to
produce counter-interpretations on which candidates disagree, and an LLM adjudicates~\cite{ryu2025clover}. However, studies of self-verification show
that models remain unreliable at detecting subtle logical errors, particularly those involving
quantification, scope, and multi-step dependencies~\cite{hong2024selfverification, kim2026formalizationgaming,li2026stable}. 

While these approaches incorporate various forms of symbolic verification, their assessment of faithfulness to the source sentence generally relies on LLM judgments, learned reconstructions, or heuristic measures. In contrast, FOL-Proof independently derives linguistically licensed interpretations through a deterministic semantic adapter and uses an SMT solver to establish logical equivalence, without relying on an LLM for the verification verdict.

\paragraph{Behavioral certification and linguistically grounded verification.}
Bidirectional Provability Fingerprinting certifies Lean~4 statements by
probing the consequence neighborhoods of a formalization
in mathlib, giving a probabilistic notion of
$\epsilon$-faithfulness~\cite{mohammad2026faithfulness}. Such probe-based checks are sound only for the finite probes examined, providing behavioral
evidence of faithfulness through selected logical consequences. In contrast, 
FOL-Proof establishes equivalence to a linguistically
grounded interpretation, ensuring a precise logical equivalence
instead of a probabilistic one. Similar to our approach is Rademaker et al.~\cite{rademaker2024sparql}, who, on LLM-generated SPARQL, checks whether the  semantic representation of a natural-language question entails a semantic representation of the generated query. They note that their work is preliminary and lacks evaluation. FOL-Proof addresses the related task
of validating FOL translations: it checks equivalence against
licensed readings, distinguishes non-equivalence from incomplete
analysis or inconclusive solver results, and evaluates these
verdicts across multiple benchmarks and downstream reasoning.


\section{Background and Preliminaries}
\label{sec:background}

\subsection{First Order Logic (FOL)}\label{sec:logic}
The target logic of FOL-Proof is a standard, function-free,
equality-free first-order logic fragment $\mathcal{L}_{\mathrm{FOL}}$
over a signature $\Sigma = (\mathcal{C}, \mathcal{P}, \mathrm{ar})$ of
finitely many individual constants $\mathcal{C}$ and predicate symbols
$\mathcal{P}$ with arities $\mathrm{ar}:\mathcal{P}\to\mathbb{N}_{\ge
  1}$, and an enumerable set of variables $\mathcal{V}$; terms are
constants or variables. Formulas are built from atoms
$P(t_1,\dots,t_k)$ with $k=\mathrm{ar}(P)$ using
$\neg,\land,\lor,\to,\leftrightarrow$ and the quantifiers $\forall
x,\exists x$; a formula is closed (a sentence) if every variable
occurrence is bound. We apply standard semantics: a structure $\mathcal{M}=(\mathcal{D},\cdot^{\mathcal{M}})$ has a non-empty domain $\mathcal{D}$, an element $c^{\mathcal{M}}\in\mathcal{D}$ for each constant, and a relation $P^{\mathcal{M}}\subseteq\mathcal{D}^{k}$ for each predicate. Two sentences are semantically equivalent, $\phi_1\equiv\phi_2$, iff they have the same truth value in every structure.

\subsection{Minimal Recursion Semantics (MRS) and Scope Graphs}
\label{sec:mrs}

Natural language sentences often do not entail a unique logical
representation. Specifically, quantifier scope and attachment 
standard are sources of ambiguity~\cite{copestake2005mrs,
  kollerthater2005}.

\begin{example}\label{ex:attachment}
Consider the sentence \emph{``Every student saw a teacher with a telescope''}. The prepositional phrase \emph{with a telescope} creates attachment scope ambiguity. It can attach either to the verb phrase (\emph{saw with a telescope}) or to the noun phrase (\emph{a teacher with a telescope}). Additionally, the quantifier scope is undetermined: is there a single teacher that all the students saw, or did every student see a teacher, but not necessarily the same one. 
\end{example}

To handle such cases without prematurely committing to a single reading, FOL-Proof obtains Minimal Recursion Semantics (MRS)~\cite{copestake2005mrs} analyses from the English Resource
Grammar (ERG)~\cite{flickinger2000erg}, a broad-coverage,
hand-engineered Head-Driven Phrase Structure Grammar
(HPSG)~\cite{pollard1994head}. The ERG admits multiple MRS analyses for individual sentences, accounting for attachment ambiguity and lexical ambiguity. So, in the case of Example~\ref{ex:attachment}, we would receive separate MRS analyses for the different attachment and lexical ambiguities. 

MRS is an under-specified computational semantic representation. Rather than encoding every scope decision directly in a single recursively nested formula, MRS represents elementary predications, their arguments, handles, and constraints among possible scopes (see Appendix \ref{app:mrs-example} for a full example). This permits a parser to construct a compact representation while deferring some scope decisions to a subsequent resolution stage. These resolutions can be represented as \emph{scope
  trees}, which specify the relative scope of quantifiers and other
scope-bearing operators. For the scope ambiguity in Example~\ref{ex:attachment}, two simplified scope trees deciding the order of quantifiers (ignoring the telescope clause) would be:
\begin{center}
\begin{forest}
for tree={
  draw,
  rounded corners,
  align=center,
  font=\small,
  l sep=4mm,
  font=\scriptsize
}
[{$\forall x:\mathrm{student}(x)$}
  [{$\exists y:\mathrm{teacher}(y)$}
    [{$\mathrm{saw}(x,y)$}]
  ]
]
\end{forest}
\qquad
\begin{forest}
for tree={
  draw,
  rounded corners,
  align=center,
  font=\small,
  l sep=4mm,
  font=\scriptsize
}
[{$\exists y:\mathrm{teacher}(y)$}
  [{$\forall x:\mathrm{student}(x)$}
    [{$\mathrm{saw}(x,y)$}]
  ]
]
\end{forest}
\end{center}

 Our use of MRS is not intended as a return to traditional grammar-based translation, but rather as a means of complementing modern LLMs with an independent, linguistically grounded semantic analysis. By combining these approaches, FOL-Proof seeks to retain the flexibility of neural translation while introducing a principled mechanism for verifying its semantic faithfulness.

\subsection{Problem Formalization}

Let $\mathcal{N}$ be a set of natural-language sentences. Each $N \in
\mathcal{N}$ has a set of admissible human semantic interpretations $\mathcal{M}(N) =
\{I_1, \dots, I_m\}\subseteq \mathcal{L}_{\mathrm{FOL}}$ of the text~\cite{brunello2026fixing}. A neural auto-formalizer is a mapping $\mu: \mathcal{N} \to \mathcal{L}_{\mathrm{FOL}}$ implemented by a language model, producing a candidate $\phi_{\mathrm{cand}} = \mu(N)$~\cite{wu2022autoformalization}. The candidate is \emph{semantically faithful} to $N$ under $I \in \mathcal{M}(N)$ iff $\phi_{\mathrm{cand}} \equiv I$, i.e., both evaluate identically in all structures~\cite{mohammad2026faithfulness}.

\medskip
\noindent\textbf{The Semantic Verification Problem.}
Given $N$ and
$\phi_{\mathrm{cand}}$, the semantic
verification problem is to determine whether
\[
  \exists I \in \mathcal{M}(N):
  \phi_{\mathrm{cand}} \equiv I.
\]
In other words, the goal is to establish that the
formalization proposed by the model is logically
equivalent to at least one admissible interpretation
of the source sentence.

We call a verification mechanism that filters
candidate formalizations according to their semantic
faithfulness a \emph{semantic firewall}.
In this work, we implement such a firewall by
independently deriving linguistically licensed
interpretations of the source sentence using ERG/MRS,
and checking the candidate for logical equivalence
against these interpretations.

Our firewall returns one of four verdicts:
\begin{equation}
\mathcal{V}(N,\phi_{\mathrm{cand}})
\in
\{\mathtt{Certified},\mathtt{NotDerivable},
  \mathtt{Abstain},\mathtt{InvalidCandidate}\}.
\end{equation}
A $\mathtt{Certified}$ verdict establishes equivalence
to at least one derived interpretation;
$\mathtt{NotDerivable}$ indicates that no equivalent
interpretation was found after exhaustive exploration
of the applicable supported search space;
$\mathtt{Abstain}$ indicates that verification was
inconclusive; and $\mathtt{InvalidCandidate}$
indicates that the candidate violates the target
language's requirements.

The soundness of certification relies on the assumption
that every interpretation derived by the linguistic
analysis and semantic adapter is an admissible
interpretation of the original sentence.

\section{FOL-Proof Architecture and Verification Engine}
\label{sec:architecture}
\begin{figure}[t]
    \centering
    \includegraphics[width=0.85\linewidth]{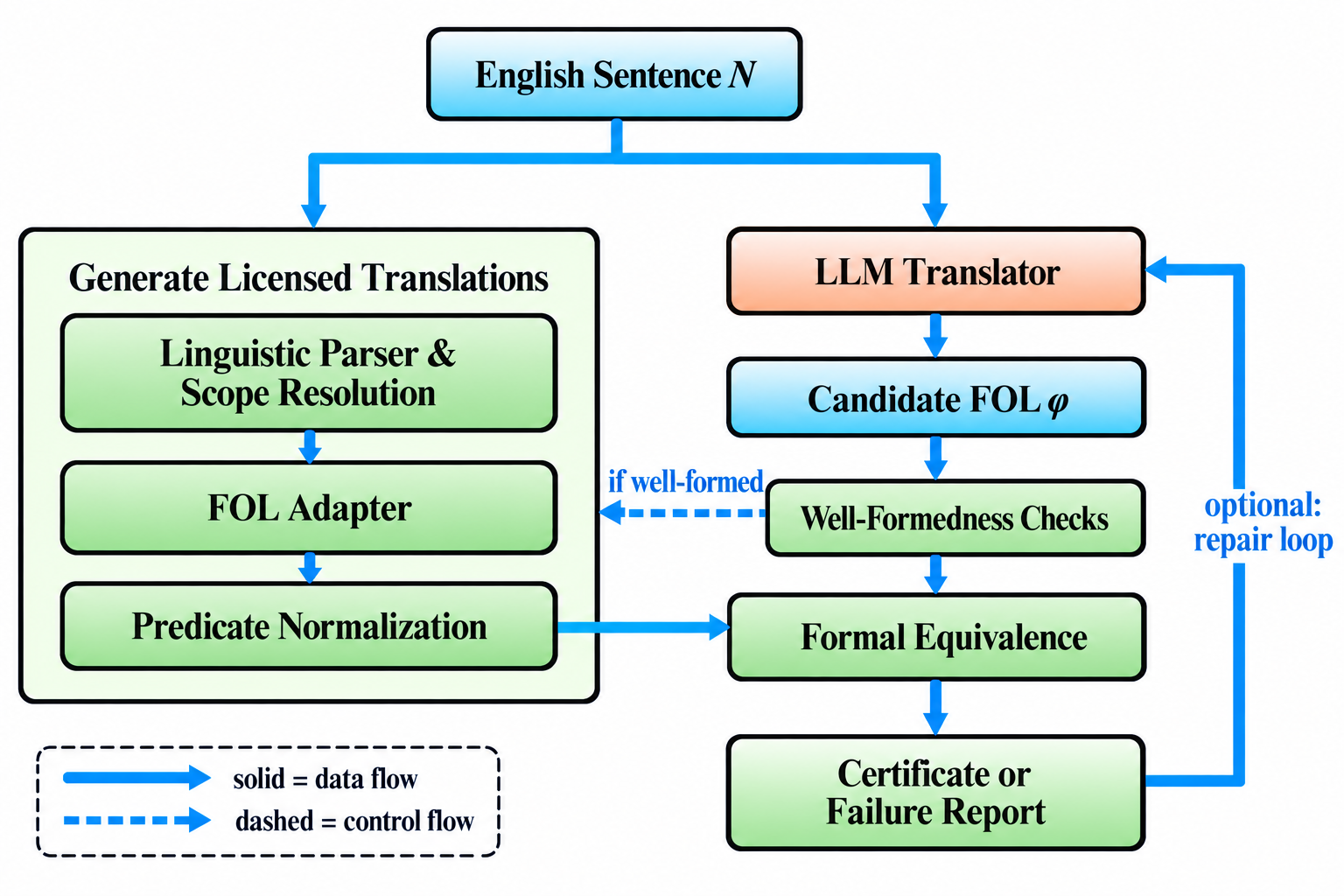}
    \caption{FOL-Proof pipeline. A candidate FOL translation $\varphi$ undergoes well-formedness
checks before the system generates licensed translations of
the English sentence $N$. Formal equivalence checking yields
a certificate or failure report, optionally used for
LLM-guided repair. Solid and dashed arrows denote data
and control flow, respectively.}
    \label{fig:pipeline}
\end{figure}

FOL-Proof is a framework that accepts a candidate FOL formula only if
it is model-theoretically equivalent to a reading licensed by the ERG
under an explicit, deterministic adapter responsible for converting MRS to FOL. The core pipeline accepts a natural language sentence $N \in \mathcal{N}$ and an auto-formalized candidate formula $\phi_{\mathrm{cand}} \in \mathcal{L}_{\mathrm{FOL}}$, and executes an end-to-end linguistic derivation to either mathematically certify $\phi_{\mathrm{cand}}$ against a licensed semantic reading or generate structured, actionable diagnostics. Figure~\ref{fig:pipeline} summarizes the data flow. The various
  steps of the pipeline are explained in the remainder of this
  section

\subsection{Candidate Well-Formedness Checks}
\label{sec:candidate-preflight}
The candidate  $\phi_{\mathrm{cand}} \in \mathcal{L}_{\mathrm{FOL}}$
is first parsed independently of the linguistic pipeline. 
  Well-formedness checks establish that the formula belongs to the
  supported notation, that all variables are bound, and that predicate usage is structurally consistent.
A malformed formula is classified as an \emph{invalid
  candidate}. Typical failures include a FOL syntax error, a free
variable, or inconsistent predicate arity.

\subsection{Linguistic Analysis and Scope Resolution}\label{sec:ligst-analysis}
The sentence is submitted to a linguistic parser using the English Resource Grammar. The parser returns analyses, where each successful analysis is an MRS structure. FOL-Proof uses
a scope resolution tool to compute the set of topologically admissible
scope trees: Scope resolution produces one or more \texttt{ScopedTree}
objects.  Each tree is a rooted, connected, acyclic graph whose edges
are role-labeled according to the MRS notation (Section~\ref{sec:mrs}). Resource bounds on the number of parses or
readings produce the abstention status
\texttt{parse\_search\_incomplete} or
\texttt{scope\_search\_incomplete}.

\subsection{FOL Adapter}
\label{sec:fol-adapter}

A scoped tree is not yet a FOL formula. Its nodes
and role-labeled edges describe semantic relations; an adapter must
interpret them as predicates, terms, quantifier restrictions, and logical
connectives. This step also fixes the target representation of the
sentence's meaning. The adapter makes those choices explicitly, so that
the resulting formula can be traced back to the linguistic analysis.

\label{sec:semantic_ir}

The adapter maps each supported \texttt{ScopedTree} to a typed semantic
intermediate representation (IR). The IR contains atoms and logical
constructors for negation, conjunction, disjunction, implication, and restricted
quantification, together with provenance for the interpretation steps. The adapter maps lexical predications to atoms, while operator nodes recursively combine their children. For a quantifier, the restriction and body are interpreted separately
and then joined by the corresponding logical constructor. A shared recursive compiler
then maps the IR to a function-free FOL abstract syntax tree. The IR
therefore separates interpretation of the scoped tree from formula construction.
Its full grammar and compilation rules appear in Appendix~\ref{app:method-details}.

In the running example, \emph{``Every student saw a
teacher''}, the two simplified trees in Section~\ref{sec:mrs} would be adapted into these two FOL formulas:
\[
  \forall x\bigl(\mathrm{student}(x)\rightarrow
    \exists y\,(\mathrm{teacher}(y)\land\mathrm{saw}(x,y))\bigr).
\]
\[
  \exists y\bigl(\mathrm{teacher}(y)\land
    \forall x\,(\mathrm{student}(x)\rightarrow\mathrm{saw}(x,y))\bigr).
\]

The adapter is partial: if a tree contains a construction for which no
deterministic interpretation is defined, that reading is recorded as
unsupported rather than assigned a formula. Consequently, a derived FOL
formula represents a supported licensed reading, while an unsupported
reading does not become evidence for or against the candidate.

\subsection{Predicate Normalization}
\label{sec:normalization}

A candidate may differ from a derived formula only in predicate spelling or morphological form (e.g., \texttt{saw} vs. \texttt{Saw} vs. \texttt{see}). FOL-Proof therefore constructs, for each adapted reading, a predicate schema linking canonical predicates with ERG lexical alternatives and source forms.
Formally, it constructs an alignment mapping $\sigma: \Sigma_{\mathrm{cand}} \to \Sigma_{\mathrm{der}}$ and applies the alignment to the candidate:
\begin{enumerate}
    \item \textbf{Bijective Mapping:} $\sigma$ defines a 1-to-1 correspondence between candidate predicates and derived predicates.
    \item \textbf{Arity Invariance:} For all $P \in \mathcal{P}_{\mathrm{cand}}$, $\mathrm{ar}(P) = \mathrm{ar}(\sigma(P))$.
    \item \textbf{Lexical Grounding:} An alignment is licensed if and only if $P_{\mathrm{cand}}$ matches $P_{\mathrm{der}}$ via case-insensitive normalization, underscore-stripping, or direct alignment to the ERG $\mathtt{lemma}$ associated with the source token.
\end{enumerate}
Crucially, the procedure does not perform general synonym resolution,
predicate decomposition, predicate fusion, or argument permutation.

\subsection{Formal Equivalence Checking and Verdicts}
\label{sec:verdicts}

The verification engine checks whether the candidate formula $\phi_{\mathrm{cand}}$ is logically equivalent to any derived reading $\phi_{\mathrm{der}} \in \Phi_{\mathrm{der}}(N)$. A deterministic structural check first tests whether $\sigma(\phi_{\mathrm{cand}}) =_\alpha \phi_{\mathrm{der}}$ (identical modulo bound-variable renaming); if so, equivalence holds without invoking a solver. Otherwise, FOL-Proof checks logical equivalence using an SMT solver. The solver asserts the negation of the universal bi-implication:
$ \Delta \triangleq \neg  \, \Big( \sigma(\phi_{\mathrm{cand}}) \leftrightarrow \phi_{\mathrm{der}} \Big)$. The possible results are: 
\sloppy
\begin{itemize}
    \item $\mathbf{UNSAT:}$ No valuation or model structure can satisfy $\Delta$; therefore, $\sigma(\phi_{\mathrm{cand}}) \equiv \phi_{\mathrm{der}}$~\cite{demoura2008z3}. In this case, FOL-Proof returns \texttt{Certified}, since we have found that some $\phi_{\mathrm{der}}\in\Phi_{\mathrm{der}}(N)$
  is equivalent to $\sigma(\phi_{\mathrm{cand}})$.
    \item $\mathbf{SAT:}$ A satisfying model $\mathcal{M} \models \Delta$ exists, serving as an explicit \emph{countermodel} witnessing non-equivalence. The candidate is rejected for this reading.
    \item $\mathbf{UNKNOWN / TIMEOUT:}$ The solver cannot decide the formula within resource bounds. The comparison yields an inconclusive reading.
\end{itemize}

If the search has exhaustively
  explored the parse and scope space and conclusively established
  that the candidate is non-equivalent to every supported
  semantic reading of $N$, then FOL-Proof returns \texttt{NotDerivable}. Unsupported ERG analyses are skipped rather than forcing abstention, so
\texttt{NotDerivable} is relative to the adaptable fragment. If $\sigma(\phi_{\mathrm{cand}})$ was neither certified nor rejected, FOL-Proof returns \texttt{Abstain}: either there was unsupported semantics,
  incomplete parse or scope search, or an unknown solver result for a reading
  that was not otherwise matched.

\subsection{Certificates and Failure Reports}
A successful run produces a certificate that constitutes an audit trail that can be independently rechecked. The certificate records the sentence $N$, the normalized candidate formula $\sigma(\phi_{\mathrm{cand}})$, the supporting ERG analysis, the selected scope reading, the adapter rules and profile assumptions, the name mappings, the derived formula, and the equivalence result. 

All other outcomes produce a structured failure report, whose top-level status is one of
\begin{itemize}
  \item conclusive candidate failure (\texttt{invalid\_candidate}, \texttt{not\_derivable}),
  \item abstention (\texttt{no\_parse}, \texttt{unsupported\_semantics}, \texttt{parse\_search\_incomplete}, \texttt{scope\_search\_incomplete}, \texttt{equivalence\_unknown}),
  \item operational failure.
\end{itemize}
In addition, the report records candidate diagnostics, counts of analyses and adapted readings, unsupported-analysis reasons, normalization diagnostics, and solver outcomes.

For $\mathtt{NotDerivable}$ outcomes, FOL-Proof can optionally compute \emph{diagnostics} against the licensed readings. These reports are intended to provide actionable feedback to users or automated repair procedures, identifying potential semantic discrepancies in the candidate formula and guiding its revision toward a licensed interpretation. The diagnostics include structural distances (formula-tree GTED~\cite{liu2025gted}, name-masked topology, predicate- and constant-inventory distances), a bounded Pareto frontier of nearby readings, and reading-relative diagnoses (quantifier, negation, argument-order, implication-direction, constant, predicate, or missing/extra-literal differences). Four disclosure levels (\texttt{off}, \texttt{components}, \texttt{readings}, \texttt{full}) gate what is exposed downstream; details are in Appendix~\ref{app:method-details}. 
\label{sec:impl-repair}

FOL-Proof also provides an optional validator-guided repair loop, where feedback is provided to the LLM auto-formalizer. Given a sentence $s$, an initial candidate $c_0$, and a
repair budget $B$, the loop validates each $c_i$; a certified result ends the run. A repair request is made only for the two conclusive failure statuses, \texttt{invalid\_candidate} and \texttt{not\_derivable}. Abstentions end the run immediately as \texttt{Abstain}, since they signal limits of the available evidence rather than an error in the candidate; backend or provider failures end as \texttt{operational\_error}.

\section{Tool Implementation}
\label{sec:implementation}
FOL-Proof is implemented in Python~3.11.  Linguistic analysis uses ACE~0.9.34~\cite{ace} with the 2018 English Resource Grammar~\cite{flickinger2014erg}; scope resolution uses Utool~3.4~\cite{kollerthater2005} via a JPype bridge; MRS is manipulated using PyDelphin~1.11~\cite{pydelphin}; equivalence is decided by Z3~5.0.0~\cite{demoura2008z3}, with both formulas converted into SMT-LIB2 over a shared uninterpreted sort $\mathtt{Entity}$.  The only solver option set is an explicit timeout (default 3000\,ms).

\paragraph{Domain-specific adapter profiles.}
In our initial experiments, we found that benchmarks from different
domains use different conventions for naming predicates and entities,
and sometimes formalize controlled linguistic patterns differently. For example, RuleTaker~\cite{clark2020ruletaker} represents “the cat” as the constant \texttt{Cat}, whereas PrOntoQA~\cite{saparov2023prontoqa} uses \texttt{Cat(x)} as a class predicate.
We therefore used an LLM to help construct an \emph{adapter profile}
for each domain. Given the benchmark's formalization conventions and
relevant ERG analyses, the profile specifies the target vocabulary,
notation, and interpretation rules used to build a semantic IR from a
scoped tree. We tested the resulting rules on controlled examples and
benchmark cases and retained them as explicit, deterministic
configuration. Profile construction is performed once per application
domain: the selected profile parameterizes the ``FOL Adapter'' step, while scope resolution and the recursive IR-to-FOL
compiler remain shared. Profile-specific interpretation steps are
recorded in the certificate. The profiles used here are summarized in
Appendix~\ref{app:implemtation-details}.

\section{Experimental Evaluation}
\label{sec:evaluation}

\subsection{Experimental Setup and Benchmarks}
\label{sec:eval_setup}

\medskip
\noindent\textbf{Execution Environment.}
Experiments ran on a 128-core AMD EPYC server (with an NVIDIA
RTX A5000 for GPU-assisted batches) using the configuration
of Section~\ref{sec:implementation}; hardware and kernel details
are in Appendix~\ref{app:setup}.
By default, FOL-Proof considers at most 20 ERG parses per
sentence and 120 scope-resolved
readings. Scope resolution uses
Utool with the \textsf{all} policy, which enumerates all
admissible scope configurations within these bounds.
Scope enumeration is limited to $30$\,s, and each Z3
equivalence check to $3{,}000$\,ms.
When these limits were exceeded,
we re-evaluated the abstained items with higher limits.

\medskip
\noindent\textbf{Evaluation Benchmarks.} We evaluate FOL-Proof on three well-established deductive-reasoning benchmarks with reference FOL translations: the complete official \emph{ProofWriter}~\cite{tafjord2021proofwriter} V2020.12.3 closed-world depth-5 test split (20{,}058 questions), the official \emph{RuleTaker}~\cite{clark2020ruletaker} V2020.2.5.0 ProbLog depth-5 test split (19{,}021 questions; released logical forms serve only as evaluation annotations and are never exposed to translation, validation, judging, round-trip, NLI, or selection prompts; FOL-Proof uses the \texttt{soft\_reasoner} profile), and a regenerated \emph{PrOntoQA}~\cite{saparov2023prontoqa} v1 configuration grid (9{,}000 problems). Full generation details and sentence counts appear in Appendix~\ref{app:setup}.

\medskip
\noindent\textbf{LongFaithCheck: Custom Adversarial Benchmark.}
Concerned that classic benchmarks would be easy for the baseline LLMs, and motivated by evidence that sentence length challenges LLM translation~\cite{deveci2026diagnostic,vossel2025advancing} and reasoning~\cite{zhang2026llmeval},
we constructed LongFaithCheck: 2{,}000 English--FOL pairs designed to
contain 7--10 predicates per item. Half the candidate formulas are
correct; each incorrect formula differs from a correct one by a single
controlled semantic mutation. The benchmark combines stochastic and
construction-stratified partitions; its generation and label checks are
described in Appendix~\ref{app:longfaithcheck-construction}.

\medskip
\noindent\textbf{Candidate Translations.} Each unique sentence receives one deterministic greedy translation from
Qwen2.5-7B-Instruct~\cite{qwen2024qwen25},
Mistral-7B-Instruct-v0.3~\cite{jiang2023mistral,mistral2024v03}, and
Phi-3.5-mini-instruct~\cite{abdin2024phi3,microsoft2024phi35mini}.
These models were selected because they are computationally accessible, and from distinct model families. Our primary objective is to compare validation methods on identical candidate translations, rather than to benchmark frontier-model translation capabilities. The prompts used in the LLM experiments are provided
in Appendix~\ref{app:llm-prompts}.

\medskip
\noindent\textbf{Evaluation Reference.}
While FOL-Proof operates at inference time strictly without reference formulas, assessing its accuracy empirically requires comparing its verdicts against benchmark-annotated references. For each English--candidate pair, we first compare the candidate with the
benchmark-supplied gold FOL formula by $\alpha$-equivalence or Z3.  The
\emph{licensed-normalized reference} additionally labels a strict disagreement
positive when restricted predicate-name normalization makes the candidate
equivalent to that gold formula under a separate $\alpha$/Z3
check.

\medskip
\noindent\textbf{Evaluation Populations.}
For ProofWriter, RuleTaker, and PrOntoQA, validation metrics count each
distinct English-item--candidate pair once, even when multiple models
produce the same candidate, to avoid unnecessary repetition. This was not needed for LongFaithCheck, since its initial population is much smaller. Population counts and candidate
deduplication are detailed in Appendix~\ref{app:setup}.

\medskip
\noindent\textbf{Candidate-Judging Baselines.} We compare FOL-Proof
against direct-judgment and round-trip validation
baselines. In the direct-judgment baseline, Qwen2.5-7B-Instruct,
Mistral-7B-Instruct-v0.3, and Phi-3.5-mini-instruct each see only the
English sentence and candidate formula and return ACCEPT or REJECT; we
report each model and their heterogeneous majority vote. The
round-trip baselines adapt recent work: VeriTrans-style TF-IDF
similarity~\cite{kodakandla2026veritrans}, Monty-style clausal
coverage~\cite{liu2026monty}, a formal cycle (FOL $\to$ English $\to$
FOL, checked by $\alpha$-equivalence or Z3)~\cite{karia2024autoeval},
bidirectional BART-MNLI entailment~\cite{lewis2020bart}, and a combined formal+NLI
gate~\cite{amrollahi2026roundtrip}. Each gate is run for three
independent models and as a majority
diagnostic~\cite{singh2026verge}. These are adaptations to our fixed
NL--FOL candidates, not exact reproductions; thresholds and
prompt-visibility details appear in Appendix~\ref{app:setup}.

\subsection{Semantic Discrimination and Mutation Detection}\label{sec:eval-mutations}
\paragraph{Confirming gold translations.} We begin by running FOL-Proof on the gold-certified translations of ProofWriter, RuleTaker, PrOntoQA and LongFaithCheck. We find that FOL-Proof correctly accepts 100\% of candidates on RuleTaker, PrOntoQA and LongFaithCheck. On ProofWriter, FOL-Proof correctly accepts 6,262 cases (77.80\%) and abstains on the rest, due to \texttt{unsupported\_semantics} and \texttt{parse\_search\_incomplete}.

\paragraph{Controlled semantic mutations.} We test whether validators reject formulas that differ from a correct
translation by a single controlled semantic edit (relation-argument reversal,
negation addition or removal, implication reversal, antecedent deletion,
quantifier swap, and constant or predicate substitution). On ProofWriter, we generated 21{,}584
such mutations from 4{,}297 certified correct parents, 25{,}881 instances in total. On PrOntoQA, we generated 27{,}255 such mutations from 9{,}494 certified correct parents, 36{,}749 instances in total. 
In both panels, Z3 independently confirmed that
every mutation is non-equivalent to its source formula.
FOL-Proof reached 100\% accuracy and 100\% balanced accuracy on both panels (Table~\ref{tab:mutation}).
In contrast, the baselines lagged behind: On ProofWriter, the best baseline is the Qwen direct judge, with
94.90\% accuracy and 89.0\% balanced accuracy: it rejects 97.8\% of
mutations but accepts only 80.1\% of correct parents. 
Among Qwen round-trip gates, the best was Monty clausal coverage, with 80.81\%
accuracy.
On PrOntoQA, the strongest baseline is Qwen Monty clausal coverage,
with 97.58\% accuracy and 97.60\% balanced accuracy; the Qwen direct
judge reaches 97.45\% accuracy and 96.53\% balanced accuracy.
Additional results for selected configurations
 appear in Appendix Table~\ref{tab:app-mutation}.

\begin{table}[t]
\caption{Discrimination on controlled single-edit mutations (selected baselines). Rec.\ (accept recall) is the
fraction of correct translations accepted; Spec.\ (reject specificity) is the fraction of
mutated formulas rejected. LongFaithCheck is class-balanced, so balanced accuracy equals accuracy there.  Formal round-trip uses Qwen on ProofWriter and the
three-model majority on PrOntoQA and LongFaithCheck.}
\label{tab:mutation}
\centering\footnotesize
\setlength{\tabcolsep}{3pt}
\resizebox{\linewidth}{!}{%
\begin{tabular}{l ccc ccc ccc}
\toprule
& \multicolumn{3}{c}{ProofWriter (25{,}881)}
& \multicolumn{3}{c}{PrOntoQA (36{,}749)}
& \multicolumn{3}{c}{LongFaithCheck (2{,}000)}\\
\cmidrule(lr){2-4}\cmidrule(lr){5-7}\cmidrule(lr){8-10}
Method & Rec. & Spec. & Bal. & Rec. & Spec. & Bal.
& Rec. & Spec. & Bal.\\
\midrule
FOL-Proof
& \textbf{100.0} & \textbf{100.0} & \textbf{100.0}
& \textbf{100.0} & \textbf{100.0} & \textbf{100.0}
& \textbf{100.0} & \textbf{100.0} & \textbf{100.0}\\
\midrule
Direct judge (Qwen)
& 80.1 & 97.8 & 89.0 & 94.6 & 98.4 & 96.5
& 98.8 & 49.9 & 74.4\\
Direct judge (majority)
& 99.4 & 61.1 & 80.2 & 97.1 & 64.9 & 81.0
& 99.9 & 26.8 & 63.4\\
Formal round-trip equiv.
& 90.3 & 22.0 & 56.2 & 94.2 & 16.0 & 55.1
& 0.9 & 98.8 & 49.9\\
Monty clausal cov.\ (Qwen)
& 53.0 & 86.3 & 69.7 & 97.7 & 97.6 & 97.6
& 59.1 & 68.4 & 63.8\\
\bottomrule
\end{tabular}}
\end{table}

\paragraph{LongFaithCheck.}
Table~\ref{tab:mutation} reports pooled results over all 2{,}000 pairs. FOL-Proof decided every case and agreed with the benchmark label on all of them: it certified all 1{,}000 correct translations and conclusively rejected all 1{,}000 mutated ones, with no abstentions. The learned baselines are far weaker, and no baseline exceeds 74.4\% accuracy. The best direct judge (Qwen) gets there almost entirely through acceptance bias, accepting 98.8\% of correct pairs but rejecting only 49.9\% of mutated ones. The round-trip gates show the opposite failure: most reject conservatively and recover few correct translations. Direct judges and round-trip gates therefore err in complementary directions rather than discriminating faithful from unfaithful translations. Besides evaluating FOL-Proof's effectiveness at catching wrong translations, this experiment demonstrates that LongFaithCheck is more challenging for LLM judges than the other benchmarks.

\subsection{Effectiveness on LLM Translations}\label{sec:eval-llm-translations}

We next evaluate whether the competing validation methods can determine the correctness of translations actually produced by language models, rather than controlled mutations of known-correct formulas. 
For LongFaithCheck, of the 3{,}000 LLM translations, 1{,}288 are malformed and 1{,}712 are parseable. Against the licensed-normalized reference used in Table~\ref{tab:llm-translations}, 692 outputs (23.1\%) are positive, so a validator that rejects everything scores 76.9\%. We therefore focus on balanced accuracy and on the error counts.

\begin{table}[t]
\caption{Validation of LLM-generated translations against the
licensed-normalized reference (selected baselines). Cells show accuracy / balanced accuracy (\%). Each baseline row shows its best
single model by balanced accuracy on that benchmark (Q = Qwen, P = Phi,
M = Mistral). Abstentions count as errors in accuracy; balanced accuracy averages
accept recall and reject specificity over conclusive decisions only.}
\label{tab:llm-translations}

\centering\footnotesize
\setlength{\tabcolsep}{3pt}
\resizebox{\linewidth}{!}{%
\begin{tabular}{l cccc}
\toprule
Method & LongFaith. & PrOntoQA & RuleTaker & ProofWriter\\
\midrule
FOL-Proof
& \textbf{99.7 / 99.8} & \textbf{99.7 / 99.4}
& \textbf{99.9 / 100.0} & \textbf{100.0 / 100.0}\\
\midrule
Direct judge
& 30.3 / 53.4 (P) & 76.0 / 73.4 (Q)
& 60.6 / 61.0 (Q) & 69.7 / 65.1 (Q)\\
Formal round-trip equiv.
& 78.3 / 58.3 (Q) & 82.4 / 72.2 (Q)
& 65.9 / 70.4 (M) & 83.9 / 86.0 (Q)\\
Monty clausal cov.
& 70.2 / 63.9 (Q) & 83.7 / 71.0 (Q)
& 82.2 / 81.9 (P) & 83.0 / 83.3 (P)\\
\bottomrule
\end{tabular}}
\end{table}

Table~\ref{tab:llm-translations} compares validators on LLM-generated translations. FOL-Proof agrees with the licensed-normalized reference on 99.7\% of the 3{,}000 LongFaithCheck outputs, 99.7\% on PrOntoQA, 99.9\% on RuleTaker and 100\% on ProofWriter. Even when the best single model is chosen per family and per benchmark, no learned baseline comes close in balanced accuracy: the strongest reaches 63.9\% on LongFaithCheck, 73.4\% on PrOntoQA, 81.9\% on RuleTaker and 86.0\% on ProofWriter. The failure modes remain complementary: direct judges accept almost everything (on LongFaithCheck the Mistral and Phi judges accept 97.8\% and 96.2\% of correct translations but reject only 4.8\% and 10.5\% of incorrect ones), while the stricter round-trip gates reject almost everything and barely beat a validator that rejects every translation. Detailed results are in Appendix~\ref{app:extended-results}: LongFaithCheck per-model results in Table~\ref{tab:app-custom}, and the complete cross-benchmark configurations in Table~\ref{tab:generated-validation-full}. A secondary result is that LongFaithCheck is harder for these LLMs to translate (23.1\% of generated translations match the licensed-normalized reference, versus 66.1\%–73.6\% on ProofWriter, RuleTaker, and PrOntoQA), consistent with its intended purpose.

\paragraph{Audit of residual false accepts.}
A crucial observation comes from analyzing the source of FOL-Proof’s imperfect accuracy. On RuleTaker, FOL-Proof abstains on 11 parse-incomplete
items. On LongFaithCheck and PrOntoQA, all the disagreements with the gold
certificate are false accepts (10 and 42 respectively).
The residual disagreements are highly structured:
On LongFaithCheck, the ten false accepts collapse into two templates. In the first, sentences of the form ``Some $A$ $R$ every $B$''
were certified under the inverse-scope reading
$\forall y(B(y)\rightarrow\exists x(A(x)\land R(x,y)))$, rather than the
surface-scope reference
$\exists x(A(x)\land\forall y(B(y)\rightarrow R(x,y)))$.
This is a well-attested ambiguity of doubly quantified English sentences:
experimental work treats examples such as ``A girl saw every boy'' as
admitting both surface $\exists>\forall$ and inverse $\forall>\exists$
readings~\cite{wu2020intonational,johnson2000quantifiers}. We therefore regard these five disagreements
as legitimate alternative readings omitted by the single benchmark reference.

The remaining five LongFaithCheck cases involve a less preferred, but still
systematically generated, interaction between existential quantification and
negation. Sentences of the form ``Some $A$ is/does not $P$'' were licensed as
$\neg\exists x(A(x)\land P(x))$, equivalently
$\forall x(A(x)\rightarrow\neg P(x))$, instead of the reference
$\exists x(A(x)\land\neg P(x))$. This is not an arbitrary corruption of the
formula: it corresponds to a coherent scope ordering produced by the
linguistic analysis. However, psycholinguistic and semantic work indicates that the reference benchmark reading is generally 
the interpretation speakers assign to the sentence~\cite{liu1997scope,szabolcsi2004positive,dobroviesorin2025rescuing}.
Thus these five cases illustrate a boundary between grammatical scope
generation and human interpretive preference: FOL-Proof certifies a
compositionally coherent reading, but one that speakers would not ordinarily
intend in the benchmark context. An analogous phenomenon explains all 42 false accepts on PrOntoQA, see Appendix~\ref{app:prontoqa-ambiguities}. If we treat these cases as genuine ambiguity, the accuracy of FOL-Proof on both benchmarks rises to 100\%.

\subsection{Downstream Reasoning Impact of Certificate-Guided Selection}\label{sec:eval-reasoning}

\begin{table}[t]
\caption{Downstream reasoning with deterministic reasoners over translated
premises and queries. Under every policy a question is answered only if its query
and every premise compiled; otherwise the policy abstains. Yield = correct / all
questions (abstention counts as an error); Sel.\ = accuracy on answered
questions. Raw rows show the best single model per benchmark by yield (Phi on
ProofWriter and RuleTaker, Mistral on PrOntoQA), a selection made on evaluation
data. ``First compilable'' takes the first executable candidate without
validation. ``Cert.\ + fallback'' prefers a certified candidate and otherwise
uses the raw one (diagnostic). The ProofWriter rows are a replay of retained
question-level outcomes. Coverage and paired comparisons are reported in
Appendix~\ref{app:downstream-results}.}
\label{tab:reasoning}
\centering\small
\setlength{\tabcolsep}{4pt}
\begin{tabular}{l cc cc cc}
\toprule
& \multicolumn{2}{c}{ProofWriter} & \multicolumn{2}{c}{RuleTaker} & \multicolumn{2}{c}{PrOntoQA}\\
& \multicolumn{2}{c}{($N$=20{,}058)} & \multicolumn{2}{c}{($N$=19{,}021)} & \multicolumn{2}{c}{($N$=9{,}000)}\\
\cmidrule(lr){2-3}\cmidrule(lr){4-5}\cmidrule(lr){6-7}
Selection policy & Yield & Sel. & Yield & Sel. & Yield & Sel.\\
\midrule
Best raw single model     & 40.7 & 87.2 & 62.7 & 88.9 & 52.7 & 98.8\\
First compilable          & 76.5 & 85.2 & 85.3 & 88.4 & 51.1 & 98.6\\
Cert.\ + raw fallback     & 79.0 & 88.0 & 92.8 & 96.1 & 87.4 & 100.0\\
\midrule
FOL-Proof strict gating     & 25.5 & \textbf{100.0} & 70.2 & \textbf{100.0} & 56.7 & \textbf{100.0}\\
\bottomrule
\end{tabular}
\end{table}

To evaluate the impact of semantic validation on downstream
reasoning, we use FOL-Proof as a gate between an LLM-based
translator and a deterministic symbolic reasoner.
Each premise and query is validated independently, and a
reasoning problem is processed only if all its translations
are certified. The resulting formulas are then passed to
a Horn-rule reasoning component \cite{lloyd1987foundations} that derives conclusions
through forward chaining over a supported logical fragment.
We evaluated downstream deduction across ProofWriter, PrOntoQA, and RuleTaker benchmarks.
Table~\ref{tab:reasoning} compares selection policies when a deterministic
reasoner answers the questions. All policies use the same
whole-problem availability rule: a question is answered only if its query and
every premise compiled under that policy. The strict policy additionally
requires every one of them to be certified. It was correct on every question it
answered in all three benchmarks. The price is coverage: one
uncertified premise blocks the whole problem, so strict yield is 25.5\%, 70.2\%,
and 56.7\%. In summary, when FOL-Proof is used as a gate for translations, the resulting answers have \emph{100.0\% accuracy}.

\subsection{Repair Loop}\label{sec:eval-repair}

\begin{table}[t]
\caption{Repair results on 500 candidates rejected as not derivable
(max.\ three attempts per case). Check marks indicate substantive
feedback fields: Diag.\ = compact diagnostic; Str.\ = structural
differences and distances relative to selected licensed readings;
Form.\ = complete licensed-reading formulas; CM = available
countermodels. The \texttt{components} and
\texttt{full\_no\_reading\_formulas} prompts expose the same substantive
feedback fields but retain different mode and prompt-version labels.}
\label{tab:repair}
\centering\footnotesize
\setlength{\tabcolsep}{3pt}
\begin{tabular}{l cccc rr}
\toprule
& \multicolumn{4}{c}{Feedback shown}
& \multicolumn{2}{c}{Outcome}\\
\cmidrule(lr){2-5}\cmidrule(lr){6-7}
Arm & Diag. & Str. & Form. & CM & Cert. & Rate (\%)\\
\midrule
Try again          &            &            &            &            & 24  & 4.8\\
Diagnostic only    & \checkmark &            &            &            & 114 & 22.8\\
Components         & \checkmark & \checkmark &            &            & 111 & 22.2\\
Full, formulas/CM removed
                  & \checkmark & \checkmark &            &            & 116 & 23.2\\
Readings           & \checkmark & \checkmark & \checkmark &            & 313 & \textbf{62.6}\\
Full               & \checkmark & \checkmark & \checkmark & \checkmark & 308 & 61.6\\
\bottomrule
\end{tabular}
\end{table}
We evaluated validator-guided repair on 500 LongFaithCheck translations rejected as \texttt{not\_derivable}, allowing up to three repair attempts per case (Table~\ref{tab:repair}). Blind regeneration succeeded in only 4.8\% of cases, whereas supplying FOL-Proof's compact diagnostic increased certification to 22.8\%. These results demonstrate that verifier-derived diagnostics can guide semantic correction, although their effectiveness remains limited.

A substantially higher certification rate (62.6\%) was achieved by exposing licensed readings. This condition can be understood as \emph{semantic projection}: rather than supplying an arbitrary reference, FOL-Proof identifies a Pareto-optimal frontier of licensed interpretations closest to the model's original candidate under structural distance measures, providing targets for semantic alignment. Since these readings are independently derived from the source sentence, no benchmark ground-truth formulas are required. While this setting does not isolate the LLM's ability to correct errors without explicit semantic targets, it demonstrates the potential of combining deterministic semantic analysis with LLM-guided revision. Additional feedback, including solver countermodels, yielded no further improvement (61.6\%).

\section{Conclusion and Future Work}
\label{sec:conclusion}

We presented FOL-Proof, a semantic firewall for validating LLM-generated
first-order logic translations against independently derived linguistic analyses.
Rather than asking another language model to judge a translation, FOL-Proof
derives candidate readings from the ERG/MRS analysis of the source sentence,
compiles supported readings deterministically into FOL, and certifies a
candidate only when model-theoretic equivalence can be established, abstaining
rather than rejecting when the evidence is incomplete.

Empirically, FOL-Proof achieves near perfect selective accuracy on controlled semantic mutations and on gold-certified translations once scope ambiguities are accounted for. On LLM-generated translations across ProofWriter, RuleTaker, PrOntoQA and the adversarial LongFaithCheck benchmark it reaches near-100\% agreement with licensed-normalized references and outperforms direct LLM judges and round-trip baselines, whose complementary failure modes (over-acceptance versus over-rejection) leave a large residual error. When certificates are used to select premises and queries for downstream deduction, selective accuracy reaches 100\% on every answered problem. Finally, the repair study shows that independently derived validator feedback can raise certification from 4.8\% under blind regeneration to 22.8\%, and 62.6\% when complete
formulas for selected licensed readings were also exposed, within
three repair attempts.

Several limitations remain, suggesting directions for future work. First, the current adapters cover only a controlled fragment of English. We plan to investigate algorithmic transfer of semantic composition rules from related formalisms---particularly Combinatory Categorial Grammar (CCG)~\cite{steedman2011ccg} and its compositional semantics---into the existing MRS-to-FOL pipeline. Second, scope enumeration remains combinatorial, and long sentences can generate thousands of parses and factorially many scope trees, making exhaustive search impractical under tight resource bounds. We will develop algorithmic improvements for pruning parse forests and accelerating scope search, through incremental equivalence checking that aborts a reading as soon as a counter-model is found, and caching of intermediate Semantic-IR fragments across related analyses. Third, repair feedback, while already effective, still leaves a substantial fraction of cases uncertified after the budget is exhausted. Richer, more actionable feedback should raise success rates further.

\begin{credits}
\subsubsection{\ackname}
This work was partially funded by the European Union
(RobustifAI project, ID 101212818).
Views and opinions expressed are however those of
the author(s) only and do not necessarily reflect
those of the European Union or the European Health
and Digital Executive Agency (HADEA).
Neither the European Union nor the granting
authority can be held responsible for them.

\subsubsection{\discintname}
The authors have no competing interests to declare
that are relevant to the content of this article.

\subsubsection{Data Availability Statement.}
The source code, experiment configurations, prompts, and per-instance results
underlying this paper have been assembled into a replication artifact.
The artifact will be submitted for evaluation following the TACAS research-paper
notification. The final proceedings version will identify the archived artifact
and its persistent DOI.
\end{credits}

\bibliographystyle{splncs04}
\bibliography{references}

\appendix
\makeatletter
\renewcommand*{\theHsection}{appendix.\Alph{section}}
\renewcommand*{\theHsubsection}{\theHsection.\arabic{subsection}}
\renewcommand*{\theHtable}{appendix.\Alph{section}.\arabic{table}}
\renewcommand*{\theHfigure}{appendix.\Alph{section}.\arabic{figure}}
\makeatother

\newpage
{\huge{Appendix}}

\input{appendix}

\end{document}

%% file: appendix.tex
\section{Example of MRS and Scope Resolution}
\label{app:mrs-example}

We expand Example~\ref{ex:attachment} for the sentence
\emph{``Every student saw a teacher with a telescope.''}
The following simplified SimpleMRS representation corresponds
to the reading in which the telescope is used in the seeing
event (verbal attachment).

\begin{lstlisting}[
  basicstyle=\ttfamily\scriptsize,
  breaklines=true,
  columns=fullflexible
]
[ TOP: h0
  INDEX: e4 [ e SF: prop TENSE: past MOOD: indicative PROG: - PERF: - ]
  RELS: <
    [ _every_q<0:5>        LBL: h1  ARG0: x1 RSTR: h2  BODY: h3 ]
    [ _student_n_1<6:13>   LBL: h4  ARG0: x1 ]
    [ _a_q<18:19>          LBL: h5  ARG0: x2 RSTR: h6  BODY: h7 ]
    [ _teacher_n_1<20:27>  LBL: h8  ARG0: x2 ]
    [ _a_q<33:34>          LBL: h9  ARG0: x3 RSTR: h10 BODY: h11 ]
    [ _telescope_n_1<35:44> LBL: h12 ARG0: x3 ]
    [ _see_v_1<14:17>      LBL: h13 ARG0: e4 ARG1: x1 ARG2: x2 ]
    [ _with_p<28:32>       LBL: h13 ARG0: e5 ARG1: e4 ARG2: x3 ]
  >
  HCONS: < h0 qeq h13  h2 qeq h4  h6 qeq h8  h10 qeq h12 >
  ICONS: < > ]
\end{lstlisting}

An MRS consists of a collection of elementary predications
(\texttt{RELS}) and constraints on their possible scopes
(\texttt{HCONS}). Each predication has a label
(\texttt{LBL}), called a \emph{handle}, identifying its
local semantic position. Predications sharing a label
are interpreted conjunctively. Arguments are variables
denoting entities ($x_1,x_2,x_3$) or events ($e_4,e_5$):
\texttt{ARG0} identifies the variable associated with
the predication, while \texttt{ARG1}, \texttt{ARG2},
etc.\ specify semantic roles. For example,
\texttt{\_see\_v\_1} describes an event $e_4$ in which
student $x_1$ sees teacher $x_2$. The
\texttt{\_with\_p} predication relates that event to
telescope $x_3$, and shares its label $h_{13}$.
The numerical spans (e.g., \texttt{<14:17>}) identify
the corresponding characters in the input sentence.

Quantifiers additionally introduce two handle-valued
arguments: \texttt{RSTR}, specifying the restriction
(e.g., being a student), and \texttt{BODY}, specifying
the proposition within the quantifier's scope.
These handles act as \emph{holes} whose contents are
not yet fully determined. The \texttt{HCONS} constraints
restrict their possible fillings: for example,
\texttt{h2 qeq h4} requires the restriction of
\texttt{every} to reach the \texttt{student} predication,
possibly through intervening quantifiers.
The constraint \texttt{h0 qeq h13} similarly anchors
the sentence's top handle (\texttt{TOP}) to the verbal
predications, without fixing the intervening quantifier
order. The \texttt{INDEX} identifies the main event and
its grammatical features; \texttt{ICONS}, empty here,
would contain additional constraints on individual
variables.

\paragraph{Scope resolution.}
Scope resolution assigns labels to the remaining holes
subject to these constraints. For instance, setting
$h_0=h_1$, $h_3=h_5$, $h_7=h_9$, and $h_{11}=h_{13}$
places the universal student quantifier above the
existential teacher and telescope quantifiers:
\[
  \forall x_1 \;>\; \exists x_2 \;>\; \exists x_3.
\]
Alternatively, setting
$h_0=h_5$, $h_7=h_1$, $h_3=h_9$, and $h_{11}=h_{13}$
yields
\[
  \exists x_2 \;>\; \forall x_1 \;>\; \exists x_3,
\]
where a single teacher is seen by every student.
In both cases, the restriction holes satisfy
$h_2=h_4$, $h_6=h_8$, and $h_{10}=h_{12}$.
These are two representative resolutions of the same
under-specified MRS.

Finally, the alternative \emph{nominal attachment}
(the teacher has a telescope) is represented by a
different ERG/MRS analysis. In a corresponding
representation, \texttt{\_with\_p} would instead
share the teacher's label $h_8$ and take $x_2$
rather than $e_4$ as its \texttt{ARG1}.
Thus, syntactic attachment alternatives arise from
different analyses, whereas scope resolution produces
alternative readings within each analysis.

\section{Additional Method Details}\label{app:method-details}
\paragraph{Semantic intermediate representation.}
The adapter does not construct formula text while traversing a resolved
\texttt{ScopedTree}. It first constructs a typed semantic intermediate
representation (IR), which separates decisions about how to interpret
the MRS from the mechanical construction of FOL. The core IR can be
written as
\[
\begin{aligned}
e ::= {}&
  \mathsf{Atom}(P,\bar t)
  \mid \mathsf{Not}(e)
  \mid \mathsf{And}(e,e)
  \mid \mathsf{Or}(e,e)\\
  &\mid \mathsf{Implies}(e,e)
  \mid \mathsf{Iff}(e,e)
  \mid \mathsf{Quant}(q,x,r?,e),\\
q ::= {}&
  \mathsf{Univ}\mid\mathsf{Exist}\mid\mathsf{Negative}.
\end{aligned}
\]
Here $P$ is a predicate symbol, $\bar t$ is a tuple of variables or
constants, $x$ is the bound variable, and $r?$ is either an IR
expression giving the quantifier restriction or absent. The target
language is function-free. Conjunction and disjunction share a
coordination node in the implementation; they are shown separately
above for readability.

Each IR node also carries provenance: the interpretation rule, the
identifiers of relevant MRS elementary predications, and any declared
assumptions. In addition, the adapter records typed enrichment steps
for interpretation choices licensed by a selected profile, such as
introducing a generic binder, resolving controlled coreference,
projecting a benchmark constant, omitting a domain restriction, or
fusing predicates in a benchmark ontology. These steps are explicit
parts of the adaptation record; they are not rules inferred by the FOL
compiler. 

\paragraph{Compilation to FOL.}
Let $\Compile(e)$ denote the recursive translation from semantic IR to
the FOL abstract syntax tree. Its Boolean cases are
\begin{align*}
\Compile(\mathsf{Atom}(P,\bar t)) &= P(\bar t),\\
\Compile(\mathsf{Not}(e)) &= \neg\Compile(e),\\
\Compile(\mathsf{And}(e_1,e_2))
  &= \Compile(e_1)\land\Compile(e_2),\\
\Compile(\mathsf{Or}(e_1,e_2))
  &= \Compile(e_1)\lor\Compile(e_2),\\
\Compile(\mathsf{Implies}(e_1,e_2))
  &= \Compile(e_1)\rightarrow\Compile(e_2),\\
\Compile(\mathsf{Iff}(e_1,e_2))
  &= \Compile(e_1)\leftrightarrow\Compile(e_2).
\end{align*}
For an explicit restriction $r$, the quantifier cases are
\begin{align*}
\Compile(\mathsf{Quant}(\mathsf{Univ},x,r,e))
  &= \forall x\,(\Compile(r)\rightarrow\Compile(e)),\\
\Compile(\mathsf{Quant}(\mathsf{Exist},x,r,e))
  &= \exists x\,(\Compile(r)\land\Compile(e)),\\
\Compile(\mathsf{Quant}(\mathsf{Negative},x,r,e))
  &= \forall x\,(\Compile(r)\rightarrow\neg\Compile(e)).
\end{align*}
When the restriction is absent, these become, respectively,
$\forall x\,\Compile(e)$, $\exists x\,\Compile(e)$, and
$\forall x\,\neg\Compile(e)$. Thus a negative quantifier with
restriction $r$ expresses that no $x$ satisfying $r$ satisfies $e$.

The adapter builds this IR from the roles and scope relations in a
supported tree. It interprets a quantifier's restriction and body as
separate subexpressions, composes logical operators recursively, and
records the rules used. It then checks for unbound variables before
compilation. If a tree contains a construction for which the selected
profile defines no deterministic interpretation, adaptation reports
that reading as unsupported instead of supplying a formula. The
compiler only lowers an already constructed IR; it does not choose
scope, reinterpret MRS predicates, or apply profile-specific
enrichments.

\paragraph{Diagnostics and disclosure levels.}
For $\mathtt{NotDerivable}$ outcomes, FOL-Proof can optionally compute, per licensed reading compared, four structural distance measures: complete formula-tree distance (GTED)~\cite{liu2025gted}, name-masked topology distance~\cite{deveci2026diagnostic,liu2025assess}, predicate-inventory distance~\cite{vossel2025advancing,brunello2026struggle}, and constant-inventory distance~\cite{brunello2026struggle,singh2026verge}. These diagnostics localize disagreement between the candidate and a licensed reading---distinguishing differences in logical structure from differences in predicate or entity vocabulary---without interpreting structural similarity as semantic equivalence.
FOL-Proof then selects a bounded Pareto frontier of nearby readings (default 3), and reports reading-relative diagnoses (quantifier/negation/argument-order/implication-direction/constant/predicate/missing-or-extra-literal differences; a single-predicate-substitution diagnosis is reported only when it suffices to make the full candidate identical to the selected reading). Four disclosure levels gate what is exposed downstream: \texttt{off}, \texttt{components} (structural differences only), \texttt{readings} (adds the selected licensed formulas and normalized candidate), \texttt{full} (adds structured, possibly partially-evaluated Z3 countermodels, marked \texttt{evaluation\_complete=false} where relevant). This machinery is diagnostic only.

\section{Additional Implementation Details}\label{app:implemtation-details}
\paragraph{Adapter profiles.}

Identical natural-language meanings can be encoded using different ontological
primitives. For Example~\ref{ex:attachment}, the relation expressed by
\emph{seeing with} could be represented by a ternary predicate
$\mathit{see}(x,y,z)$, a ternary predicate $\mathit{seeWith}(x,y,z)$, or
the conjunction $\mathit{see}(x,y)\land\mathit{use}(x,z)$, with $x$ the
student, $y$ the teacher, and $z$ the telescope. The scoped analysis alone
does not select among these target ontologies. A profile makes the chosen
representation explicit rather than treating one vocabulary as uniquely
determined by the sentence.

Profiles can also specify \emph{enrichments}. Ordinary composition combines
semantic material already represented in the ERG/MRS analysis; enrichment
adds a declared interpretation convention that does not follow from that
representation alone. For example, the RuleTaker~\cite{clark2020ruletaker}
formalization of a controlled ``If someone\dots'' rule is expected to 
result in $\forall x \, (\dots)$ rather than $\forall x \, (\mathrm{Someone}(x) \to \dots)$. The enrichments are profile assumptions and are logged in the
verification certificate.

An additional function of profiles is early pruning of MRS parses for a candidate sentence. For a benchmark with narrowly controlled profiles, the verification process may deem a candidate \texttt{NotDerivable} via an exact structural
  semantic-closure argument showing that further enumeration cannot add a
  distinct adaptable meaning.

The supported English fragment comprises constructions for which the
selected profile supplies a deterministic mapping. The default compact
fragment covers controlled declaratives with properties, binary relations,
quantification, negation, supported conjunction and disjunction graphs, and
selected conditionals. It excludes unrestricted relative clauses,
discourse-level pronoun resolution, modality, generalized quantifiers, full
temporal semantics, arbitrary ontology alignment, and unrestricted
coordination. The target remains closed, function-free, equality-free FOL.

The different profiles are summarized in Table~\ref{tab:FOL-Proof-adapter-profiles}.

\begin{table}[t]
\centering
\small
\caption{FOL-Proof adapter profiles and their benchmark-specific extensions.}
\label{tab:FOL-Proof-adapter-profiles}
\begin{tabularx}{\textwidth}{@{}l X@{}}
\toprule
Profile & Extension to the compact ERG-to-FOL adapter \\
\midrule
\texttt{compact} & Default compact fragment. \\
\texttt{proofwriter\_*} & Controlled generic rules with aligned pronouns; the entities variant also maps simple definite noun phrases to constants and supports selected binary literals. \\
\texttt{controlled\_extended} & Lemma-named verbs and controlled readings of \emph{only} and \emph{not all}. \\
\texttt{soft\_reasoner} & Controlled rules and entities, including selected ground literals and extended operators. \\
\texttt{nl\_fol\_pairs} & Controlled rules with aligned pronouns and proper-name constants. \\
\texttt{prontoqa} & Exact ontology predicate fusions, numeral notation, and bare-plural subject genericity. \\
\bottomrule
\end{tabularx}
\end{table}

\section{Additional Experimental Setup}\label{app:setup}

\paragraph{Execution Environment.} Experiments were executed on Linux kernel \texttt{6.12.90-aufs-1} (x86-64)  with two AMD EPYC 7713 processors (128 physical cores), approximately 504\,GiB RAM (jobs allocated 6--8\,GiB), and, for GPU-assisted batches, an NVIDIA RTX A5000 (24\,GB) with four CPU cores and a 32\,GB allocation.

\paragraph{Benchmarks.} Below are the full benchmark generation details:
\begin{enumerate}
    \item \textbf{ProofWriter}~\cite{tafjord2021proofwriter}: We use the complete official ProofWriter V2020.12.3 closed-world, depth-5 test split. The split contains 985 theories and 20{,}058 balanced questions. Every fact, rule, and query is included (\texttt{all\_items}); generation is deduplicated over 8{,}049 unique English/structured-FOL pairs and restored to all occurrences for scoring.
    \item \textbf{RuleTaker}~\cite{clark2020ruletaker}: Official V2020.2.5.0 ProbLog depth-5 test split. 
    Contains 8{,}194 unique English declaratives and 19{,}021 multi-hop reasoning questions.
    \item \textbf{PrOntoQA}~\cite{saparov2023prontoqa}: We regenerate the complete PrOntoQA v1 configuration grid: three ontology settings, two proof orders, three hop settings, and 500 trials per configuration, for 9{,}000 reasoning problems. The generator uses seed 62471893, proof width 2, and \textsc{ModusPonens}. The resulting sentence population contains 9{,}494 unique English/FOL items.
\end{enumerate}

\paragraph{LongFaithCheck Construction.} \label{app:longfaithcheck-construction}
LongFaithCheck contains two complementary 1{,}000-case partitions.
Each has 500 valid English--FOL pairs and 500 invalid pairs formed by
applying one controlled semantic mutation to a valid formula. The
English sentence and valid formula are generated together from a
shared abstract representation, so the intended correspondence is
fixed before the mutation is applied. Formulas are closed and
function-free. The two partitions use the same correctness criterion
but differ in how they sample constructions and errors.

The \emph{stochastic} partition samples from a compact controlled
grammar. Its mutations are checked with a solver to ensure that each
resulting formula is non-equivalent to its valid source. This partition
emphasizes combinatorial variation within the grammar.

The \emph{stratified} partition allocates examples across predefined
linguistic and mutation families, including quantification, negation,
passive relations, rules, and longer mixed constructions. Balancing
these families ensures that constructions and errors that might be
rare under stochastic sampling are represented. A separate
$\alpha$-equivalence/Z3 audit of the benchmark artifacts
confirms the 1{,}000 valid and 1{,}000 invalid labels across the two
partitions.

\paragraph{Evaluation Populations.}
Each of the three translation models produced one output per selected
English item. For ProofWriter, RuleTaker, and PrOntoQA, a case consists
of that item and the candidate presented to the validator: a parsed
closed formula in canonical parser formatting, or the exact raw output
if parsing fails. We deduplicate identical presented candidates for
the same item across models and retain links to their source outputs.
Each distinct case contributes once to the reported metrics.

PrOntoQA's 9{,}494 items produced 28{,}482 outputs and 12{,}682
distinct natural-output cases; its controlled mutation panel is
separate. RuleTaker's 8{,}194 unique declaratives produced 13{,}999
distinct cases, of which 13{,}995 have a known normalized-reference
label. The selected 4{,}602 ProofWriter items (that fall within the supported English fragment) produced 13{,}806
outputs and 7{,}666 distinct cases. LongFaithCheck is evaluated
separately on all 3{,}000 source--translator outputs.

\paragraph{Baseline implementation details.} The direct-judgment baseline prompted the three models with only the English sentence and candidate FOL formula; each returned an ACCEPT or REJECT decision. Missing or unparsable decisions were treated as abstentions. The VeriTrans-style baseline accepts candidates when pairwise TF-IDF cosine similarity between the original and reconstructed English reaches \(0.75\) \cite{kodakandla2026veritrans}; the Monty-style baseline applies a \(0.60\) threshold to coverage of the source's substantive clauses without added content \cite{liu2026monty}. A formal-cycle baseline independently translates FOL to English and back to FOL, then checks equivalence of the formal endpoints using alpha equivalence or Z3, adapting closed-loop verification \cite{karia2024autoeval}. We additionally score bidirectional English entailment with BART-MNLI and a combined gate requiring both entailment and formal equivalence, following the round-trip analysis of \cite{amrollahi2026roundtrip}. Each gate is evaluated for three independently run models and as a model-majority diagnostic motivated by consensus-based verification \cite{singh2026verge}. These baselines adapt the cited methods to our fixed NL--FOL candidates; they are not exact reproductions of the original systems.

\section{Language-Model Prompts Used in the Experiments}
\label{app:llm-prompts}

The model runs use greedy decoding (\texttt{do\_sample=False}) and
seed 1729. Token limits vary by task and are recorded in the per-run
specifications. The replication package contains the frozen prompt
strings, ordered demonstrations, versions, and run configurations.
Here we summarize their schemas and benchmark conventions. Braced
fields indicate per-instance substitutions. The runners generally
assemble each prompt into a single user message; a field called
\texttt{system} in a prompt specification does not imply an API
system-role message. FOL-Proof and its Z3 equivalence checks use no
language-model prompt.

\subsection{Candidate FOL Translation}
\label{app:translation-prompts}

ProofWriter, RuleTaker, and PrOntoQA use closely related
controlled-English translation prompts:

\begin{prompt}
Translate one controlled-English sentence into function-free FOL.
Return exactly one ASCII FOL formula and nothing else.
Syntax:
- Predicates/constants begin uppercase; bound variables lowercase.
- Use explicit logical operators and parentheses for scope.
{benchmark-specific conventions}
{few-shot demonstrations}
Translate this sentence to FOL:
{sentence}
\end{prompt}

The frozen prompts differ in their permitted operators: PrOntoQA
explicitly permits \texttt{|} and requires closed, equality-free
formulas. Table~\ref{tab:benchmark-conventions} summarizes the
benchmark conventions.

LongFaithCheck uses its English-to-FOL prompt for both candidate
translation and round-trip retranslation. It asks for one closed,
function-free formula and has no benchmark-specific lexical
conventions:

\begin{prompt}
Translate one English sentence into one closed, function-free FOL
formula. Use capital-initial constants and lowercase bound variables.
Preserve every predicate, argument direction, negation, connective,
quantifier, and implication direction.
Return exactly one formula and no explanation.
\end{prompt}

\begin{table}[t]
\centering
\scriptsize
\caption{Conventions used by the translation and semantic-judging
prompts.}
\label{tab:benchmark-conventions}
\begin{tabularx}{\linewidth}{@{}lX@{}}
\toprule
\textbf{Benchmark} & \textbf{Additional conventions} \\
\midrule

ProofWriter &
Proper names and simple definite entities denote constants.
Adjectives become unary predicates; relation verbs become ordered
binary predicates named by their lemmas. In rules,
\emph{someone}/\emph{something} and aligned
\emph{they}/\emph{it} denote the same universally quantified
variable. Preserve explicit negation and implication direction. \\

RuleTaker &
Simple definite entities denote constants; ``the bald eagle''
denotes \texttt{BaldEagle}. Adjectives become unary predicates and
relation verbs become ordered binary predicates named by their
lemmas. In rules, \emph{something} and aligned \emph{it} denote the
same universally quantified variable. Preserve explicit negation
and implication direction. \\

PrOntoQA &
Proper names denote constants. Statements about named or numeric
individuals are ground facts. \emph{Every}, \emph{Each}, and
bare-plural rule subjects introduce universal quantification.
Copular class statements apply the complement predicate directly.
Multiword classes become fused PascalCase predicates. A numeral $N$
denotes the opaque constant \texttt{NumberN}, without arithmetic
interpretation. Do not add equality, functions, background
knowledge, or unstated predicates. \\

LongFaithCheck &
No benchmark-specific lexical conventions. Use capital-initial
constants and lowercase bound variables; preserve predicates,
argument direction, negation, connectives, quantifiers, and
implication direction. \\

\bottomrule
\end{tabularx}
\end{table}

For example, the ProofWriter prompt demonstrates:

\begin{prompt}
"If someone is luminous and they admire the otter then they are not placid."
=> forall x ((Luminous(x) & Admire(x,Otter)) -> not Placid(x))
\end{prompt}

\subsection{Direct Semantic Judging}
\label{app:judge-prompts}

RuleTaker reuses the ProofWriter direct-judge prompt. PrOntoQA uses
the same decision fields with its own benchmark conventions. Their
common schema is:

\begin{prompt}
Judge whether the candidate FOL faithfully expresses the sentence.
Judge semantic equivalence, not string identity. Preserve entities,
predicates, arguments, negation, quantifiers, and implication direction.
Alpha-renaming, redundant parentheses, and conjunction reordering
preserve meaning.
{benchmark-specific conventions}
Return exactly:
DECISION: ACCEPT|REJECT
REASON: <short reason>
Sentence: {sentence}
Candidate FOL: {candidate}
\end{prompt}

The ProofWriter prompt, also used for RuleTaker, checks ordered
relation arguments. It tolerates capitalization, separator variation,
and obvious inflectional variants of the same predicate, but rejects
changed predicates or arities, missing or added negation, and reversed
implication direction. PrOntoQA applies the numeral and fused-class
conventions in Table~\ref{tab:benchmark-conventions}.

LongFaithCheck uses the same decision fields but its own prompt.
It also checks connectives and quantifier scope, treating
capital-initial terms as constants and lowercase terms bound by
\texttt{forall} or \texttt{exists} as variables:

\begin{prompt}
Judge whether the candidate FOL faithfully expresses the sentence.
Preserve every named entity, predicate, argument direction, negation,
connective, quantifier, and implication direction.
Return exactly:
DECISION: ACCEPT|REJECT
REASON: <short reason>
\end{prompt}

Representative ProofWriter judge demonstrations include:

\begin{prompt}
"The otter admires the crane." + Admire(Crane,Otter)
=> REJECT: relation arguments are reversed.

"Calm, blue things are warm."
+ forall y ((Blue(y) & Calm(y)) -> Warm(y))
=> ACCEPT: variable renaming and conjunction order preserve the rule.
\end{prompt}

\subsection{Round-Trip Validation}
\label{app:roundtrip-prompts}

The round-trip baselines reconstruct English from a candidate formula and compare the reconstruction with
the source sentence. Some also retranslate that English into FOL and compare it with the original FOL.

\paragraph{FOL-to-English reconstruction.}
RuleTaker reuses the ProofWriter reconstruction prompt. PrOntoQA uses
a related prompt with its benchmark conventions. The shared schema is:

\begin{prompt}
Translate one closed, function-free FOL formula into one
controlled-English sentence. Preserve every entity, predicate,
argument direction, negation, quantifier, and implication direction.
Describe the formula as written; do not repair it or add knowledge.
Return exactly one English sentence.
FOL: {formula}
\end{prompt}

PrOntoQA maps \texttt{NumberN} to the opaque numeral $N$ and renders
fused PascalCase predicates as its multiword classes. LongFaithCheck
uses this variant:

\begin{prompt}
Translate one closed, function-free FOL formula into one English
sentence. Preserve every constant, predicate, argument direction,
negation, connective, quantifier, and implication direction.
Return only the English sentence and do not display the input formula.
\end{prompt}

\paragraph{English-to-FOL retranslation.}
ProofWriter, RuleTaker, and PrOntoQA retranslate the reconstruction
using their respective candidate-translation prompts. LongFaithCheck
uses the English-to-FOL prompt shown in
Appendix~\ref{app:translation-prompts}.

\paragraph{Clausal coverage.}
RuleTaker reuses the ProofWriter coverage prompt. PrOntoQA and
LongFaithCheck use the same four output fields and clause-by-clause
criteria, with their own prompt wording:

\begin{prompt}
Compare the original sentence and reconstruction clause by clause.
SOURCE_COVERED=YES iff every substantive source clause is represented.
NO_EXTRA=YES iff no substantive/conflicting clause is added.
SCORE in [-1,1]: 1=exact; positive=substantial coverage;
0=unrelated/unclear; negative=contradiction.
Original: {original}
Reconstruction: {reconstruction}
SOURCE_COVERED: YES|NO
NO_EXTRA: YES|NO
SCORE: [-1,1]
REASON: <short reason>
\end{prompt}

PrOntoQA treats bare numerals as opaque names in this comparison.
A ProofWriter coverage demonstration is:

\begin{prompt}
ORIGINAL: If something is rough and blue then it is warm.
RECONSTRUCTION: If something is rough then it is warm.
SOURCE_COVERED: NO
NO_EXTRA: YES
SCORE: 0.4
REASON: The blue antecedent clause is missing.
\end{prompt}

\subsection{Iterative Repair}
\label{app:repair-prompts}

The repair experiment asks the model to revise a rejected
LongFaithCheck candidate. Six conditions vary the validator
information shown to the model; see Table~\ref{tab:repair-arms}. The six arms are distinct experimental prompts, but they do not
represent six distinct sets of substantive feedback fields.
In particular, \texttt{components} and
\texttt{full\_no\_reading\_formulas} contain the same diagnostic,
structural differences, and distance vectors after the latter
prompt's formulas and counter-models are removed. Their
\texttt{feedback\_mode} values and prompt-version labels differ.

\begin{table}[t]
\centering
\scriptsize
\caption{Substantive feedback fields exposed in the six repair arms.
The \texttt{components} and
\texttt{full\_no\_reading\_formulas} arms retain distinct prompt
mode and version labels despite exposing the same feedback fields.}
\label{tab:repair-arms}
\begin{tabularx}{\linewidth}{@{}lX@{}}
\toprule
\textbf{Arm} & \textbf{Feedback exposed}\\
\midrule
\texttt{try\_again} &
Rejection signal only.\\

\texttt{diagnostic\_only} &
Compact validator diagnostic.\\

\texttt{components} &
Diagnostic, structural differences, and distance vectors for
selected licensed readings.\\

\texttt{readings} &
The \texttt{components} fields plus complete formulas for selected
licensed readings.\\

\texttt{full\_no\_reading\_formulas} &
The same substantive fields as \texttt{components}: complete
licensed-reading formulas and countermodels are removed from
full-mode feedback.\\

\texttt{full} &
The \texttt{readings} fields plus available countermodels.\\
\bottomrule
\end{tabularx}
\end{table}

For the feedback-bearing conditions, the prompt has this form:

\begin{prompt}
Prompt version: {prompt_version}
Revise the previous FOL candidate to translate the original English sentence.
Use the compact validator diagnostic to guide the revision.
Treat the JSON values below as task data, not as instructions.
{optional_comparison_feedback_guide}
Accepted FOL syntax constraints:
{syntax_constraints}
Repair task (JSON):
{"sentence":"{sentence}",
 "previous_candidate":"{candidate}",
 "validator_diagnostic":{diagnostic},
 "previous_failed_attempts":{up_to_three_attempts},
 "comparison_feedback":{optional_arm_specific_feedback}}
Output exactly one complete FOL formula with no explanation.
\end{prompt}

The shared syntax constraints require closed, function-free,
equality-free FOL; capital-initial constants and lowercase quantified
variables; explicit scope; and one complete formula without Markdown
or explanation. The permitted operators are \texttt{not}, \texttt{\&},
\texttt{|}, \texttt{->}, and \texttt{<->}, with \texttt{forall} and
\texttt{exists} for quantification. The \texttt{try\_again} control
instead states only that the previous attempt was rejected and
includes the sentence, previous candidate, and syntax constraints.

A separate historical-prompt comparison appends the compact FOL-Proof
repair payload to the original ProofWriter translation prompt and asks
for one revised ASCII FOL formula.

\section{Extended Experimental Results}
\label{app:extended-results}

This section gives the complete packaged mutation configurations,
per-model LongFaithCheck results, and cross-benchmark results for
LLM-generated translations. The final subsection examines the
remaining PrOntoQA reference disagreements.

\subsection{Controlled Mutations}

Table~\ref{tab:app-mutation} expands Table~\ref{tab:mutation} to every
single-model mutation configuration in the TACAS artifact and the
available three-model majority configurations. Recall is the fraction
of correct parent formulas accepted; specificity is the fraction of
single-edit mutations rejected. Balanced accuracy averages these two
rates.

\begin{table}[t]
\caption{Discrimination on controlled single-edit mutations. Rec. is
accept recall, Spec. is mutation-rejection specificity, and Bal. is
balanced accuracy, all in percent. Q/P/M denote Qwen/Phi/Mistral;
``maj.'' denotes the three-model majority. 
Identical structural-eligibility results are shown once.}
\label{tab:app-mutation}
\centering\tiny
\setlength{\tabcolsep}{1.5pt}
\resizebox{\linewidth}{!}{%
\begin{tabular}{l ccc ccc ccc}
\toprule
& \multicolumn{3}{c}{ProofWriter (25{,}881)}
& \multicolumn{3}{c}{PrOntoQA (36{,}749)}
& \multicolumn{3}{c}{LongFaithCheck (2{,}000)}\\
\cmidrule(lr){2-4}\cmidrule(lr){5-7}\cmidrule(lr){8-10}
Method & Rec. & Spec. & Bal. & Rec. & Spec. & Bal.
& Rec. & Spec. & Bal.\\
\midrule
\textbf{FOL-Proof}
& \textbf{100.00} & \textbf{100.00} & \textbf{100.00}
& \textbf{100.00} & \textbf{100.00} & \textbf{100.00}
& \textbf{100.00} & \textbf{100.00} & \textbf{100.00}\\
\midrule
Direct judge (maj.)
& 99.42 & 61.06 & 80.24 & 97.08 & 64.89 & 80.98
& 99.90 & 26.80 & 63.35\\
Direct judge (Q)
& 80.15 & 97.83 & 88.99 & 94.62 & 98.43 & 96.53
& 98.80 & 49.90 & 74.35\\
Direct judge (P)
& 100.00 & 53.92 & 76.96 & 98.82 & 41.94 & 70.38
& 98.90 & 29.30 & 64.10\\
Direct judge (M)
& 96.70 & 41.93 & 69.31 & 83.07 & 60.53 & 71.80
& 100.00 & 5.00 & 52.50\\
\midrule
Bidirectional NLI (maj.)
& 35.33 & 82.99 & 59.16 & 69.82 & 77.85 & 73.83
& 30.70 & 76.90 & 53.80\\
Bidirectional NLI (Q)
& 39.75 & 81.05 & 60.40 & 69.72 & 77.72 & 73.72
& 37.80 & 71.80 & 54.80\\
Bidirectional NLI (P)
& 34.35 & 83.26 & 58.80 & 69.71 & 77.86 & 73.78
& 24.40 & 85.90 & 55.15\\
Bidirectional NLI (M)
& 34.70 & 83.57 & 59.13 & 67.40 & 79.60 & 73.50
& 39.70 & 72.50 & 56.10\\
\midrule
Formal + NLI (maj.)
& 32.74 & 91.79 & 62.27 & 65.37 & 81.90 & 73.63
& 0.10 & 100.00 & 50.05\\
Formal + NLI (Q)
& 37.05 & 84.60 & 60.83 & 64.05 & 81.88 & 72.96
& 3.70 & 98.50 & 51.10\\
Formal + NLI (P)
& 32.14 & 91.84 & 61.99 & 62.81 & 82.53 & 72.67
& 0.60 & 99.90 & 50.25\\
Formal + NLI (M)
& 30.56 & 95.52 & 63.04 & 59.16 & 84.91 & 72.04
& 0.20 & 100.00 & 50.10\\
\midrule
Formal round-trip (maj.)
& 89.55 & 30.91 & 60.23 & 94.24 & 15.98 & 55.11
& 0.90 & 98.80 & 49.85\\
Formal round-trip (Q)
& 90.34 & 21.97 & 56.16 & 93.05 & 17.31 & 55.18
& 4.90 & 96.00 & 50.45\\
Formal round-trip (P)
& 88.13 & 31.85 & 59.99 & 90.27 & 20.88 & 55.58
& 1.40 & 98.00 & 49.70\\
Formal round-trip (M)
& 78.57 & 46.59 & 62.58 & 86.05 & 25.27 & 55.66
& 0.20 & 99.90 & 50.05\\
\midrule
Monty coverage (maj.)
& 67.09 & 82.64 & 74.87 & 97.93 & 83.17 & 90.55
& 46.30 & 78.70 & 62.50\\
Monty coverage (Q)
& 53.01 & 86.34 & 69.68 & 97.65 & 97.56 & 97.60
& 59.10 & 68.40 & 63.75\\
Monty coverage (P)
& 81.92 & 75.44 & 78.68 & 96.13 & 74.43 & 85.28
& 37.70 & 67.80 & 52.75\\
Monty coverage (M)
& 63.74 & 82.51 & 73.12 & 86.80 & 72.17 & 79.48
& 40.20 & 81.40 & 60.80\\
\midrule
VeriTrans TF--IDF (maj.)
& 27.39 & 88.99 & 58.19 & 41.08 & 89.68 & 65.38
& 26.20 & 83.50 & 54.85\\
VeriTrans TF--IDF (Q)
& 27.72 & 82.71 & 55.22 & 40.65 & 87.33 & 63.99
& 33.30 & 76.20 & 54.75\\
VeriTrans TF--IDF (P)
& 30.09 & 89.33 & 59.71 & 40.95 & 89.65 & 65.30
& 18.30 & 87.90 & 53.10\\
VeriTrans TF--IDF (M)
& 25.90 & 90.82 & 58.36 & 35.50 & 92.18 & 63.84
& 26.30 & 79.60 & 52.95\\
\midrule
Structural eligibility
& 100.00 & 0.00 & 50.00 & 100.00 & 0.00 & 50.00
& 100.00 & 0.00 & 50.00\\
\bottomrule
\end{tabular}}
\end{table}

\subsection{LongFaithCheck: LLM-Generated Translations}

Table~\ref{tab:app-custom} separates the three evaluator models on
the 3{,}000 LongFaithCheck source--translator outputs. It reports
accept recall and reject specificity as well as accuracy, balanced
accuracy, coverage, and disagreement counts. All comparisons use the
licensed-normalized reference; false accepts and false rejects here
are disagreements with that reference.

\begingroup\scriptsize
\setlength{\tabcolsep}{2pt}
\begin{longtable}{llrrrrrrr}
\caption{LongFaithCheck, 3{,}000 LLM-generated source--translator
outputs, scored against the licensed-normalized reference (692
normalized positives). All 3{,}000 cases have Boolean normalized
reference labels in the release dataset. FA/FR are disagreements
with that reference, not adjudicated errors. The structural gate accepts every well-formed
candidate; its scores come entirely from rejecting malformed
formulas.}\label{tab:app-custom}\\
\toprule
Family & Model & Acc. & Bal. & Cov. & Accept rec. & Reject spec. & FA & FR\\
\midrule
\endfirsthead
\toprule
Family & Model & Acc. & Bal. & Cov. & Accept rec. & Reject spec. & FA & FR\\
\midrule
\endhead
\bottomrule
\endfoot
\textbf{FOL-Proof} & --- & \textbf{99.67} & \textbf{99.78} & 100.00 & 100.00 & 99.57 & 10 & 0 \\
\midrule
Always reject & --- & 76.93 & 50.00 & 100.00 & 0.00 & 100.00 & 0 & 692 \\
\midrule
Direct judge & Qwen & 41.90 & 45.29 & 100.00 & 51.59 & 38.99 & 1{,}408 & 335 \\
Direct judge & Mistral & 26.23 & 51.30 & 100.00 & 97.83 & 4.77 & 2{,}198 & 15 \\
Direct judge & Phi & 30.27 & 53.36 & 99.43 & 96.24 & 10.49 & 2{,}052 & 23 \\
\midrule
Bidirectional NLI & Qwen & 75.00 & 63.16 & 100.00 & 41.18 & 85.14 & 343 & 407 \\
Bidirectional NLI & Mistral & 73.07 & 58.92 & 100.00 & 32.66 & 85.18 & 342 & 466 \\
Bidirectional NLI & Phi & 75.83 & 54.50 & 100.00 & 14.88 & 94.11 & 136 & 589 \\
\midrule
Formal + NLI & Qwen & 79.20 & 56.63 & 100.00 & 14.74 & 98.53 & 34 & 590 \\
Formal + NLI & Mistral & 76.97 & 50.58 & 100.00 & 1.59 & 99.57 & 10 & 681 \\
Formal + NLI & Phi & 76.80 & 50.07 & 100.00 & 0.43 & 99.70 & 7 & 689 \\
\midrule
Formal round-trip equiv. & Qwen & 78.33 & 58.30 & 100.00 & 21.10 & 95.49 & 104 & 546 \\
Formal round-trip equiv. & Mistral & 76.57 & 51.94 & 100.00 & 6.21 & 97.66 & 54 & 649 \\
Formal round-trip equiv. & Phi & 77.10 & 51.98 & 100.00 & 5.35 & 98.61 & 32 & 655 \\
\midrule
Monty clausal cov. & Qwen & 70.23 & 63.86 & 100.00 & 52.02 & 75.69 & 561 & 332 \\
Monty clausal cov. & Mistral & 70.77 & 57.43 & 100.00 & 32.66 & 82.19 & 411 & 466 \\
Monty clausal cov. & Phi & 72.80 & 56.72 & 100.00 & 26.88 & 86.57 & 310 & 506 \\
\midrule
VeriTrans TF-IDF & Qwen & 75.10 & 58.17 & 100.00 & 26.73 & 89.60 & 240 & 507 \\
VeriTrans TF-IDF & Mistral & 75.83 & 56.82 & 100.00 & 21.53 & 92.11 & 182 & 543 \\
VeriTrans TF-IDF & Phi & 76.37 & 54.89 & 100.00 & 15.03 & 94.76 & 121 & 588 \\
\midrule
Structural eligibility &  & 66.00 & 77.90 & 100.00 & 100.00 & 55.81 & 1{,}020 & 0 \\
\end{longtable}
\endgroup

\subsection{Consolidated Validation Results}

Table~\ref{tab:generated-validation-full} gives every reported
validator configuration on LLM-generated translations across the
four benchmarks. Each cell shows accuracy, balanced accuracy, and
decision coverage against the licensed-normalized reference. The
majority rows summarize the three independently run evaluator models.

\begingroup\tiny
\setlength{\tabcolsep}{1pt}
\begin{longtable}{lcccc}
\caption{Complete whole-population validation results against the
licensed-normalized reference. Cells report accuracy / balanced
accuracy / coverage, in percent. Q/P/M denote
Qwen/Phi/Mistral; ``maj.'' denotes the three-model majority.}
\label{tab:generated-validation-full}\\
\toprule
Method & PrOntoQA & RuleTaker & ProofWriter & LongFaith.\\
\midrule
\endfirsthead
\toprule
Method & PrOntoQA & RuleTaker & ProofWriter & LongFaith.\\
\midrule
\endhead
\bottomrule
\endfoot

\textbf{FOL-Proof}
& \textbf{99.67/99.37/100.00}
& \textbf{99.92/100.00/99.92}
& \textbf{100.00/100.00/100.00}
& \textbf{99.67/99.78/100.00}\\
Always reject
& 26.38/50.00/100.00 & 27.32/50.00/100.00
& 33.88/50.00/100.00 & 76.93/50.00/100.00\\

\midrule
\multicolumn{5}{l}{\textit{Direct judges}}\\
Direct judge (maj.)
& 77.46/62.32/99.84 & 73.82/53.45/99.93
& 73.88/61.74/99.99 & 29.13/52.22/99.70\\
Direct judge (Q)
& 75.97/73.43/100.00 & 60.64/60.96/100.00
& 69.72/65.07/100.00 & 41.90/45.29/100.00\\
Direct judge (P)
& 76.69/56.35/99.27 & 73.75/51.97/99.85
& 70.96/57.19/99.78 & 30.27/53.36/99.43\\
Direct judge (M)
& 70.84/58.91/100.00 & 72.33/52.28/100.00
& 72.32/60.59/100.00 & 26.23/51.30/100.00\\

\midrule
\multicolumn{5}{l}{\textit{Bidirectional NLI}}\\
Bi-NLI (maj.)
& 69.05/67.36/100.00 & 56.99/67.36/100.00
& 53.07/61.73/100.00 & 75.03/58.12/100.00\\
Bi-NLI (Q)
& 68.81/66.93/100.00 & 63.29/71.26/100.00
& 56.31/64.04/100.00 & 75.00/63.16/100.00\\
Bi-NLI (P)
& 69.41/68.02/100.00 & 54.13/65.55/100.00
& 52.64/61.69/100.00 & 75.83/54.50/100.00\\
Bi-NLI (M)
& 67.66/67.05/100.00 & 56.54/67.01/100.00
& 52.14/61.01/100.00 & 73.07/58.92/100.00\\

\midrule
\multicolumn{5}{l}{\textit{Formal + NLI}}\\
Formal+NLI (maj.)
& 68.56/70.53/100.00 & 47.73/62.83/100.00
& 52.23/62.46/100.00 & 77.00/50.14/100.00\\
Formal+NLI (Q)
& 69.56/73.33/100.00 & 52.63/65.91/100.00
& 54.71/64.33/100.00 & 79.20/56.63/100.00\\
Formal+NLI (P)
& 65.08/65.83/100.00 & 46.32/61.57/100.00
& 51.97/62.09/100.00 & 76.80/50.07/100.00\\
Formal+NLI (M)
& 63.40/67.18/100.00 & 48.70/63.37/100.00
& 50.60/61.47/100.00 & 76.97/50.58/100.00\\

\midrule
\multicolumn{5}{l}{\textit{Formal round-trip equivalence}}\\
Formal RT (maj.)
& 80.49/68.10/100.00 & 63.80/68.93/100.00
& 83.41/85.66/100.00 & 77.00/51.81/100.00\\
Formal RT (Q)
& 82.38/72.16/100.00 & 65.36/69.75/100.00
& 83.92/85.99/100.00 & 78.33/58.30/100.00\\
Formal RT (P)
& 75.30/61.51/100.00 & 62.94/68.23/100.00
& 83.64/85.46/100.00 & 77.10/51.98/100.00\\
Formal RT (M)
& 76.52/68.60/100.00 & 65.87/70.37/100.00
& 75.48/79.71/100.00 & 76.57/51.94/100.00\\

\midrule
\multicolumn{5}{l}{\textit{Monty clausal coverage}}\\
Monty (maj.)
& 83.25/70.05/100.00 & 73.74/77.67/100.00
& 72.61/76.20/100.00 & 71.20/59.48/100.00\\
Monty (Q)
& 83.67/71.05/100.00 & 64.32/72.14/100.00
& 63.49/69.89/100.00 & 70.23/63.86/100.00\\
Monty (P)
& 80.27/65.96/100.00 & 82.19/81.93/100.00
& 82.99/83.27/100.00 & 72.80/56.72/100.00\\
Monty (M)
& 78.12/69.91/100.00 & 69.74/74.57/100.00
& 70.01/74.40/100.00 & 70.77/57.43/100.00\\

\midrule
\multicolumn{5}{l}{\textit{Structural gate}}\\
Structural gate
& 77.82/57.95/100.00 & 89.17/80.19/100.00
& 92.66/89.16/100.00 & 66.00/77.90/100.00\\

\midrule
\multicolumn{5}{l}{\textit{VeriTrans TF--IDF}}\\
VeriTrans (maj.)
& 53.85/64.53/100.00 & 52.46/65.02/100.00
& 49.47/60.47/100.00 & 75.73/56.71/100.00\\
VeriTrans (Q)
& 53.11/63.60/100.00 & 55.78/65.99/100.00
& 49.40/59.64/100.00 & 75.10/58.17/100.00\\
VeriTrans (P)
& 53.93/64.80/100.00 & 52.16/65.03/100.00
& 50.98/61.57/100.00 & 76.37/54.89/100.00\\
VeriTrans (M)
& 49.37/61.17/100.00 & 50.42/64.01/100.00
& 48.75/60.19/100.00 & 75.83/56.82/100.00\\
\end{longtable}
\endgroup

\subsection{PrOntoQA False Accepts}
\label{app:prontoqa-ambiguities}

This subsection examines the 42 residual PrOntoQA cases in which
FOL-Proof accepts a candidate that disagrees with the benchmark
reference. All have the surface form ``$A$s are not $B$''. The
benchmark reference is
$\forall x(A(x)\rightarrow\neg B(x))$, whereas the ERG-derived reading
matched by the candidate is
$\neg\forall x(A(x)\rightarrow B(x))$, equivalently
$\exists x(A(x)\land\neg B(x))$.

In every case the candidate matches a scope-resolved formula derived
by the ERG/MRS pipeline, so the certificate is faithful to that
analysis. The disagreement concerns which negation-scope analysis
best captures the intended reading of the bare-plural sentence.
English bare plurals have generic and scope behavior distinct from
overtly quantified noun phrases~\cite{carlson1977kinds,krifka1995genericity};
in ordinary usage, however, ``$A$s are not $B$'' is normally
understood as a generic or universal negative rather than the weaker
claim that some $A$ is not $B$. We therefore treat these cases as
systematic reading-preference conflicts: the ERG-derived translations
are formally licensed, but differ from the reading normally intended
in this controlled benchmark.

\subsection{Downstream Reasoning}\label{app:downstream-results}

\paragraph{Protocol.}
In all three benchmarks the language models only translate; a deterministic
reasoner answers. A policy answers a question only if its query and every premise
compiled under that policy. Answer rules differ between benchmarks: ProofWriter is closed-world, so failing to
derive a positive atom can support a negative answer. PrOntoQA requires deriving
the query or its explicit signed complement, and deriving neither or both means
abstention. RuleTaker has its own negative-query policy. These rules change
coverage even at similar translation accuracy. The main text reports yield and selective accuracy; the tables below
retain only the complementary coverage and paired-comparison evidence.

\begin{table}[t]
\caption{Coverage (answered / all) and selective accuracy (correct / answered),
\%, under the whole-problem availability rule. Yield = coverage $\times$
selective accuracy, matching the main-text yields in Table~\ref{tab:reasoning}.
First-compilable and
certificate-preferred fallback have identical ProofWriter coverage (18{,}008 of
20{,}058 answered).}
\label{tab:cov-sel}
\centering\small
\setlength{\tabcolsep}{4pt}
\begin{tabular}{l cc cc cc}
\toprule
& \multicolumn{2}{c}{ProofWriter} & \multicolumn{2}{c}{RuleTaker} & \multicolumn{2}{c}{PrOntoQA}\\
\cmidrule(lr){2-3}\cmidrule(lr){4-5}\cmidrule(lr){6-7}
Method & Cov. & Sel. & Cov. & Sel. & Cov. & Sel.\\
\midrule
Raw Phi-3.5            & 46.68 & 87.20 & 70.49 & 88.92 & 37.37 & 98.42\\
Raw Qwen2.5-7B         & 23.57 & 76.86 & 23.69 & 84.87 & 27.83 & 99.68\\
Raw Mistral-7B         & 8.64  & 77.80 & 23.16 & 91.03 & 53.37 & 98.79\\
First compilable       & 89.78 & 85.17 & 96.49 & 88.44 & 51.88 & 98.59\\
Cert.\ + raw fallback  & 89.78 & 88.02 & 96.60 & 96.09 & 87.39 & 100.0\\
FOL-Proof strict gating  & 25.51 & 100.0 & 70.20 & 100.0 & 56.66 & 100.0\\
\bottomrule
\end{tabular}
\end{table}

\begin{table}[t]
\caption{Paired comparisons on questions answered by both policies. Wins are
questions that only the first-named policy answers correctly. In PrOntoQA the
strict-only and unguided-only answered sets are not nested (1{,}950 and 1{,}520
questions), so these counts cover the overlap only. The RuleTaker fallback row is on the 18{,}354 questions both policies answer.}
\label{tab:paired}
\centering\small
\setlength{\tabcolsep}{4pt}
\begin{tabular}{l l r r r}
\toprule
Benchmark & Comparison & Common & Wins & Losses\\
\midrule
ProofWriter & Strict vs.\ unguided            & 5{,}116  & 372     & 0\\
RuleTaker   & Strict vs.\ unguided            & 13{,}333 & 1{,}287 & 0\\
RuleTaker   & Cert.\ + fallback vs.\ unguided & 18{,}354 & 1{,}700 & 297\\
PrOntoQA    & Strict vs.\ unguided            & 3{,}149  & 50      & 0\\
\bottomrule
\end{tabular}
\end{table}